\documentclass[journal]{IEEEtran}
\usepackage[T1]{fontenc}
\usepackage{color}
\usepackage{flushend}
\usepackage{amsmath}
\usepackage{cite}

\usepackage{booktabs}
\usepackage{tabu}
\usepackage[cmintegrals]{newtxmath}
\usepackage{bm}
\usepackage{indentfirst}
         
\usepackage{amsmath,amssymb}
\usepackage{array}
\usepackage{mathrsfs}
\usepackage{cite}
\usepackage{tikz}
\usepackage{color}
\usepackage{amssymb}
\usepackage{url}
\usepackage{hyperref}
\usepackage{todonotes}
\usepackage{color}
\usepackage{bbding}
\usepackage{multirow}
\usepackage{colortbl}
\usepackage{amsmath}
\usepackage{graphicx}
\usepackage{subcaption}
\usepackage{algorithm}
\usepackage{algpseudocode}
\usepackage{amsmath}

\begin{document}

\title{Propose and Rectify: A Forensics-Driven MLLM Framework for Image Manipulation Localization}

\author{Keyang~Zhang, 
Chenqi~Kong,~\IEEEmembership{Member,~IEEE}, 
Hui~Liu,
Bo~Ding,

Xinghao~Jiang,~\IEEEmembership{Senior Member,~IEEE},
Haoliang~Li,~\IEEEmembership{Member,~IEEE}
\thanks{K. Zhang, H. Liu, B. Ding, H. Li are with the Department of Electrical Engineering, City University of Hong Kong, Hong Kong SAR. (email: keyazhang4-c@my.cityu.edu.hk, liulxs99@gmail.com, bo.ding@my.cityu.edu.hk, haoliang.li@cityu.edu.hk).}
\thanks{C. Kong is with the Rapid-Rich Object Search (ROSE) Lab, School of Electrical and Electronic Engineering, Nanyang Technology University, Singapore. (email: chenqi.kong@ntu.edu.sg.)}
\thanks{X. Jiang is with the Shanghai Jiao Tong University, China. (email: xhjiang@sjtu.edu.cn.)}
}

% \thanks{H. Li is the corresponding author.}

\markboth{Submitted to IEEE Transactions on Information Forensics and Security}%
{Shell \MakeLowercase{\textit{et al.}}: Bare Demo of IEEEtran.cls for IEEE Communications Society Journals}

\maketitle

\begin{abstract}
The increasing sophistication of image manipulation techniques demands robust forensic solutions that can both reliably detect alterations and precisely localize tampered regions. Recent Multimodal Large Language Models (MLLMs) show promise by leveraging world knowledge and semantic understanding for context-aware detection, yet they struggle with perceiving subtle, low-level forensic artifacts crucial for accurate manipulation localization. This paper presents a novel Propose-Rectify framework that effectively bridges semantic reasoning with forensic-specific analysis. In the proposal stage, our approach utilizes a forensic-adapted LLaVA model to generate initial manipulation analysis and preliminary localization of suspicious regions based on semantic understanding and contextual reasoning. In the rectification stage, we introduce a Forensics Rectification Module that systematically validates and refines these initial proposals through multi-scale forensic feature analysis, integrating technical evidence from several specialized filters. Additionally, we present an Enhanced Segmentation Module that incorporates critical forensic cues into SAM's encoded image embeddings, thereby overcoming inherent semantic biases to achieve precise delineation of manipulated regions. By synergistically combining advanced multimodal reasoning with established forensic methodologies, our framework ensures that initial semantic proposals are systematically validated and enhanced through concrete technical evidence, resulting in comprehensive detection accuracy and localization precision. Extensive experimental validation demonstrates state-of-the-art performance across diverse datasets with exceptional robustness and generalization capabilities.

\end{abstract}


\IEEEpeerreviewmaketitle

\section{Introduction}

\IEEEPARstart {I}{n} an era where digital images serve as dominant media in social platforms \cite{hochman2014social}, and as primary evidence in legal proceedings \cite{mnookin1998image}, ensuring their authenticity has become increasingly critical \cite{kong2022digital}. The proliferation of sophisticated image editing software and the emergence of powerful generative artificial intelligence models \cite{brooks2023instructpix2pix} have fundamentally transformed the digital manipulation landscape. These advanced technologies now enable the creation of highly convincing synthetic content and seamlessly altered images that often evade detection by human observers, posing unprecedented challenges to the credibility of visual evidence and forensic assessment \cite{thakur2020recent}.

\begin{figure}[t]
    \centering
    \includegraphics[width=0.8\linewidth]{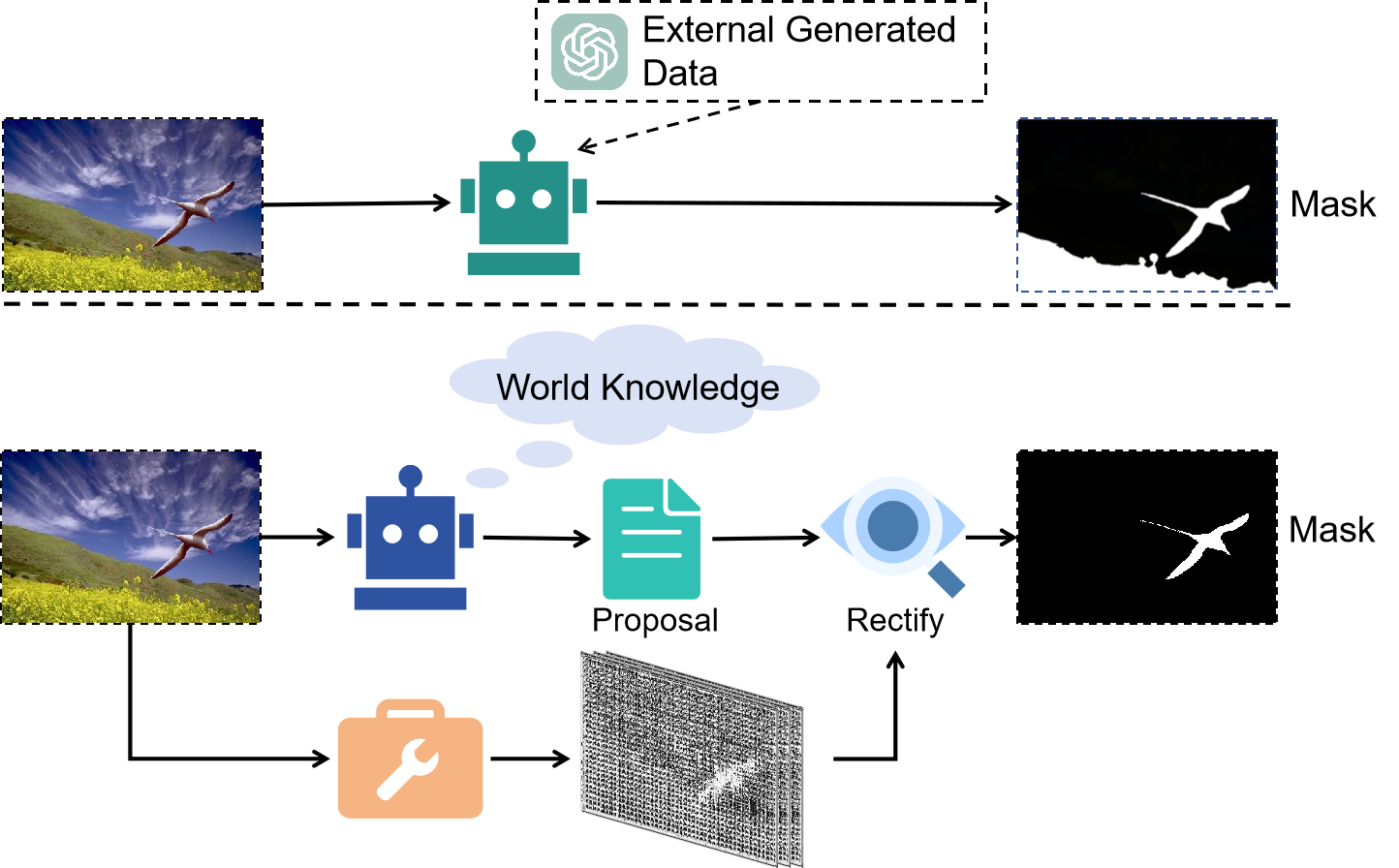}
    \caption{Comparison between previous MLLM-based methods and our proposed framework. Previous methods rely on external generated data and directly output manipulation judgments through single-pass image analysis. In contrast, our Propose-Rectify paradigm first invokes MLLMs to generate initial proposals by analyzing images with world knowledge, then systematically validates and rectifies these proposals through multiple forensic features, ensuring more robust and reliable detection results.}
    \vspace{-5mm}
    \label{fig:comparison}
\end{figure}

Previous image manipulation detection approaches have primarily relied on handcrafted feature extractors \cite{yerushalmy2011digital, cao2009accurate, kwon2022learning, zhou2018learning} or constraint learning architectures \cite{li2019localization, dong2022mvss, wu2019mantra} to analyze the image statistical characteristics \cite{xiang2024research} and identify inconsistencies \cite{kong2025pixel, kong2022detect, guillaro2023trufor, luo2010jpeg} introduced during post-processing operations such as splicing \cite{shi2007natural, cozzolino2015splicebuster}, copy-move \cite{fridrich2003detection, lin2009fast}, and inpainting \cite{wu2021iid, barglazan2024image}. While these methods have achieved reasonable success in their corresponding manipulation scenarios, they exhibit significant limitations when deployed in complex real-world applications. Specifically, they often struggle with generalization across diverse unseen manipulation techniques, suffer performance degradation under perturbations like compression and noise, and most critically, lack the capability to incorporate semantic understanding and high-level analysis that are essential for comprehensive forensic evaluation.

The recent emergence of Multimodal Large Language Models such as LLaVA \cite{liu2023visual} and Qwen-VL \cite{wang2024qwen2} has opened new avenues for advancing image forensics by fundamentally addressing these limitations through global semantic perception and sophisticated analytical capabilities. Unlike conventional approaches constrained to statistical pattern detection, recent MLLM based methods \cite{liu2024forgerygpt,xu2024fakeshield, huang2025sida} demonstrate remarkable potential by integrating extensive world knowledge with high-level contextual understanding across diverse analytical dimensions, including text recognition, lighting consistency analysis, physical law adherence, and contextual plausibility assessment \cite{yildirim2024task}. After fine-tuned on well-annotated forensic datasets, MLLM frameworks can provide interpretable explanation of detected anomalies or identify tampered region through reasoning processes independently.

Despite these promising capabilities, current MLLM-based frameworks face architectural limitations that constrain their effectiveness. Their modality encoders are inherently optimized for semantic understanding through general image-caption pairs \cite{radford2021learning}, rendering them inadequate for perceiving and utilizing the subtle, low-level forensic artifacts essential for precise manipulation localization. Furthermore, several existing methods \cite{xu2024fakeshield, liu2024forgerygpt, huang2024ffaa} exhibit over-reliance on GPT-generated textual descriptions of tampered regions for training prompt generation in mask decoding or for achieving explainability. This dependency introduces several critical vulnerabilities: the generated descriptions cannot guarantee accuracy across all analyzed aspects and may contain hallucinations, while the massive volume of data makes comprehensive, human-supervised verification both impractical and expensive. This approach also creates a risky model distillation scenario, where the forensic system's performance becomes critically dependent on the capabilities of the data generation model, such as GPT-4o \cite{achiam2023gpt}. 

% Furthermore, critical forensic traces such as noise pattern inconsistencies \cite{guillaro2023trufor}, compression artifact \cite{luo2010jpeg} irregularities, and pixel-level statistical anomalies \cite{kong2025pixel} are inherently beyond the guaranteed detection capabilities of current frameworks, which lack reliable mechanisms for perceiving such low-level features and cannot adequately represent them explicitly through text descriptions. These limitations become problematic when confronting advanced manipulations that carefully preserve semantic plausibility while only introducing low-level forensic traces that require multi-view visual examination rather than linguistic interpretation.

In actual forensic investigations, after obtaining an initial understanding of the image and preliminary identification of potentially tampered regions, forensic professionals systematically employ specialized analytical tools to extract multi-dimensional features, enabling more precise localization of manipulated areas and accurate authenticity assessments \cite{piva2013overview, nist}. This established forensic workflow raises a fundamental question: can MLLMs be guided to develop analogous capabilities, utilizing appropriate specialized tools for auxiliary analysis and localization, thereby enhancing the reliability and precision of automated forensic assessments?

Motivated by this observation, this paper presents a novel Propose-Rectify pipeline that systematically bridges high-level semantic understanding with fine-grained forensic analysis. As shown in Fig. \ref{fig:comparison}, the framework operates through a two-stage process: First, proposing potential manipulation regions and detection hypotheses using forensic-adapted MLLMs that leverage contextual clues and semantic inconsistencies to identify suspicious areas; Second, rectifying these initial assessments through rigorous validation using multiple complementary forensic features that analyze low-level artifacts, compression patterns, and statistical anomalies to confirm or refine the preliminary findings. This progressive Propose-Rectify approach, enhanced with a forensic-aware segmentation module, harnesses the strengths of multimodal reasoning while grounding decisions in concrete forensic evidence, effectively mitigating the individual weaknesses of semantic-only or pixel-level detection methods.

In this work, we introduce several key technical innovations:
\begin{itemize}
\item We establish a systematic \textbf{Propose-Rectify Pipeline} that bridges high-level semantic understanding from forensic-adapted MLLMs with precise low-level forensic analysis, effectively addressing the inherent limitation of current MLLM-based approaches in perceiving subtle manipulation artifacts.

\item We design a specialized \textbf{Forensics Rectification Module} that leverages initial MLLM analysis to intelligently select appropriate forensic features for multi-scale examination, subsequently employing these targeted features to rectify both detection results and segmentation boundaries of tampered regions with enhanced precision.

\item To further enhance the prediction accuracy, we introduce a \textbf{Enhanced Segmentation Module} that integrates critical low-level forensic cues into SAM encoded image embeddings, effectively overcoming their inherent semantic biases to ensure accurate delineation of manipulated regions, even those lacking clear semantic boundaries.

\item Extensive quantitative and qualitative experimental results consistently demonstrate that our proposed method achieves state-of-the-art performance across diverse datasets and outperforms the baselines in generalization and robustness evaluations.

\end{itemize}

The remainder of this paper is organized as follows: Sec. II reviews related work in image manipulation detection and localization, multimodal large language models, and rectification algorithms. Sec. III presents our proposed Propose-Rectify pipeline, detailing the forensic-adapted MLLM component, multi-scale rectification module, and forensic-aware enhanced segmentation module. Sec. IV provides comprehensive experimental evaluation across multiple datasets. Finally, Sec. V concludes the paper with discussion of key contributions, current limitations, and promising directions for future research.

\section{Related Work}
\subsection{Image Manipulation Detection and Localization}
Early image manipulation detection algorithms focused on exploiting statistical irregularities and camera-specific artifacts left by the natural image acquisition pipeline. These methods demonstrated that tampering operations could disrupt the correlations established during in-camera processing by analyzing Color Filter Array interpolation patterns \cite{cao2009accurate,ferrara2012image}, lens distortions \cite{fu2012forgery,gloe2010efficient,yerushalmy2011digital}, and noise characteristics \cite{kobayashi2010detecting,lyu2014exposing}. Compression-based techniques exploited inconsistencies in JPEG artifacts \cite{bianchi2012image,chen2011detecting,farid2009exposing}, while geometric approaches detected perspective \cite{yao2011detecting} and lighting anomalies \cite{peng2016optimized} in composite images. Although these handcrafted feature extraction methods provided interpretable results and computational efficiency, they generally suffered from limited accuracy when confronted with sophisticated editing tools and required extensive domain knowledge for effective deployment across diverse image types and quality levels.

The advent of deep neural networks has fundamentally transformed image forensics by enabling automatic feature learning and end-to-end optimization for manipulation detection and localization tasks. With the learned forensic representations on intrinsic traces \cite{cozzolino2019noiseprint, zhou2018learning, guillaro2023trufor, kong2025pixel} or compression patterns \cite{kwon2022learning,rao2022towards,wang2022jpeg}, the learning based methods achieve substantial improvements over traditional handcrafted methods. Modern architectures have evolved from single-view analysis to sophisticated multi-view frameworks \cite{hu2020span, dong2022mvss, wu2019mantra} that extract features from complementary perspectives and at different granularities, employing pyramid attention networks and progressive spatial-channel correlations to capture comprehensive manipulation signatures. The integration of high-frequency filters has proven particularly effective, with steganalysis rich model (SRM) \cite{wu2019mantra, zhou2018learning, rao2021self} and Bayar filters \cite{dong2022mvss, wu2019mantra} capturing abundant forgery artifacts that complement learned deep features. Boundary-aware detection has emerged as another crucial strategy, with methods specifically targeting forgery boundaries to significantly improve pixel-level localization accuracy \cite{dong2022mvss, kong2025pixel}. Transformer-based architectures represent the current frontier \cite{liu2023explicit}, with frameworks like TruFor \cite{guillaro2023trufor} concurrently fusing high-level RGB features and low-level noise traces, while PIM \cite{kong2025pixel} focuses on the inherent pixel correlations involved in the demosaicing process. Despite these advances, contemporary learning-based methods continue to face challenges in cross-dataset generalization and robustness against increasingly sophisticated manipulation techniques.

\subsection{Multimodal Large Language Models}
The past few years have witnessed transformative advancements in Large Language Models, with transformer architectures enabling unprecedented text comprehension and generation capabilities through strategic scaling. This progress has naturally extended to Multimodal Large Language Models, which integrate linguistic sophistication with perception capabilities across image, video, and audio modalities. This evolutionary leap enables cross-modal processing and reasoning, establishing foundations for visual question answering, image captioning, document understanding, and multimodal dialogue systems. Prominent MLLMs including LLaVA \cite{liu2023visual}, GPT-4 \cite{achiam2023gpt}, and Qwen-VL \cite{wang2024qwen2} leverage specialized vision encoders like CLIP \cite{radford2021learning} alongside innovative fusion mechanisms to align visual and textual representations, enabling applications that extend beyond traditional text-only paradigms into complex multimodal reasoning scenarios.

Building upon these general multimodal capabilities, MLLMs have increasingly been adapted for region-specific visual tasks that require precise spatial understanding. LISA \cite{lai2024lisa} introduces reasoning segmentation that generates binary masks from text queries demanding logical reasoning and world knowledge, demonstrating how MLLMs can bridge semantic understanding with spatial localization. Expanding on this foundation, frameworks like Seg-Zero \cite{liu2025seg} further enhance reasoning segmentation by decoupling the reasoning and segmentation processes, utilizing reinforcement learning to activate emergent test-time reasoning capabilities for improved performance.

In the forensics domain specifically, researchers have explored various approaches to leverage MLLM capabilities for tampering detection. Fakeshield \cite{xu2024fakeshield} utilizes multi-aspect text analysis of forged regions as prompts for mask decoding, demonstrating the potential of language-guided forensic analysis. Meanwhile, other works focus on achieving explainability in forensic decisions: both ForgeryGPT \cite{liu2024forgerygpt} and M2F2-Net \cite{guo2025rethinking} forward detection masks into MLLMs for interpretation, while FFAA \cite{huang2024ffaa} aligns different hypothesized answers with image authenticity and attempts to select the optimal one through systematic comparison. While these approaches have explored promising prospects for integrating MLLM's powerful reasoning with forensics tasks, they exhibit critical limitations. Their performance heavily depends on textual descriptions from other base models, and they fail to incorporate specialized forensic tools for technical validation. Consequently, they remain susceptible to hallucination problems and often fail to achieve accurate localization of tampered regions.

\subsection{Rectification Algorithms}
The paradigm of algorithmic rectification, wherein initial outputs undergo progressive correction or improvement through iterative processes or complementary information integration, has emerged as a powerful approach across diverse artificial intelligence domains. In natural language processing, rectification mechanisms enable LLMs to enhance their output quality by iteratively evaluating and refining generated responses through multi-step reasoning processes, ultimately producing more accurate and logically consistent results \cite{cheng2025empowering,wu2024get,guo2025deepseek}. Recent developments have further advanced this concept through the integration of external resources, as agent-like models attempt to invoke specialized tools \cite{huang2023agentcoder}, databases \cite{lei2024spider}, or communication systems \cite{chen2024internet} to validate and correct their intermediate reasoning steps. Such external validation has proven particularly effective in mathematical reasoning, factual verification, and complex problem-solving scenarios. This rectification philosophy has also found compelling applications in forensics area, where they progressively enhance precision by hierarchical forgery attribute detection \cite{guo2023hierarchical} or coarse-to-fine examinations \cite{jia2018coarse}. Despite the proven efficacy of refinement approaches within individual modalities, the systematic exploration of cross-modal refinement paradigms, particularly those that leverage high-level semantic understanding from multimodal large language models to guide fine-grained forensic analysis for image manipulation detection, represents a promising yet underexplored research direction.

\begin{figure*}[t]
    \centering
    \includegraphics[width=1\linewidth]{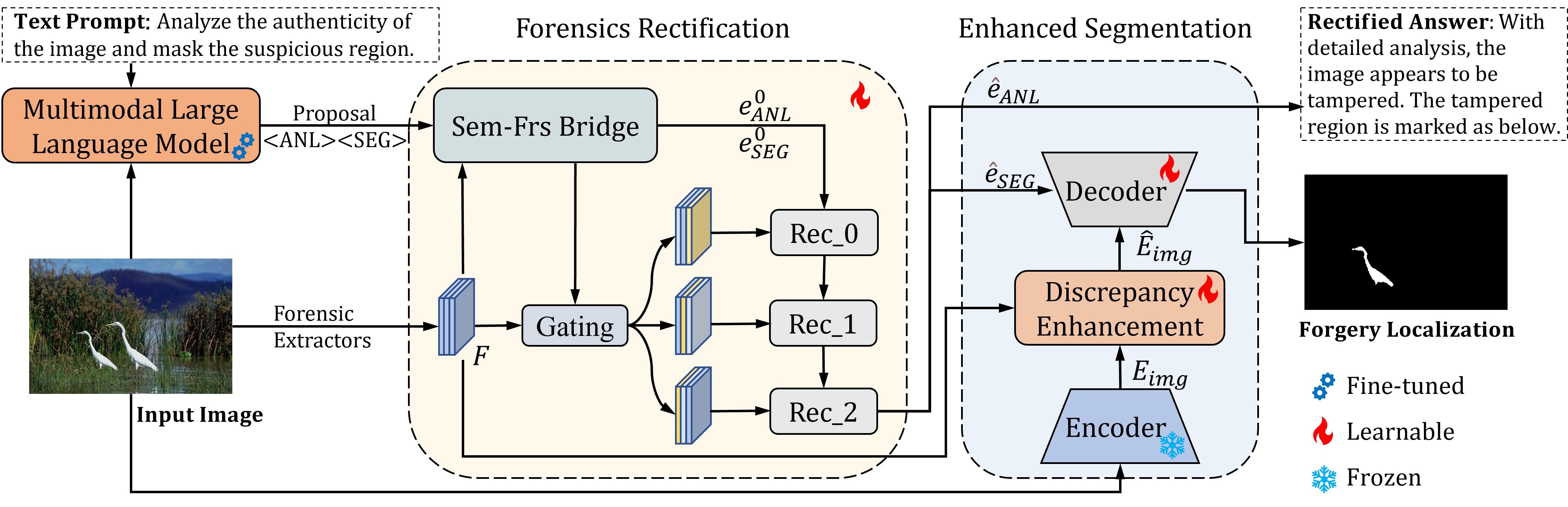}
    \caption{The overall \textbf{Propose-Rectify} framework for manipulation detection and localization. A forensic-adapted MLLM analyzes the image and generates an initial proposal while a forensics multi-feature extractor simultaneously captures a series of forensic artifacts. The Forensics Rectification Module then progressively rectifies the proposal embeddings with extracted forensic evidences through multi-scale analysis and produces rectified detection results. When tampering is detected, the Enhanced Segmentation Module amplifies semantic-forensic discrepancies in SAM encoded embeddings to enable more precise manipulation mask generation.}
    \vspace{-5mm}
    \label{fig:pipeline}
\end{figure*}

\section{Proposed Method}
This section presents the proposed manipulation detection and localization framework. We begin by introducing the overall Propose-Rectify pipeline that sequentially processes input images through initial semantic proposal generation followed by forensic evidence-based rectification for comprehensive analysis. Subsequently, we elaborate on the key components and their design principles, including the forensic-adapted MLLM for initial proposal generation, the Multi-Features Extractors for low-level artifact capture, the Forensics Rectification Module for semantic embedding refinement, and the Enhanced Segmentation Module for precise boundary delineation. Lastly, we present the end-to-end training strategy that jointly optimizes detection accuracy and localization precision across all framework components.

\subsection{Overall framework}

In this paper, we introduce our Propose-Rectify paradigm that enhances multimodal large language model proposals with low-level forensic evidence for accurate manipulation detection and localization. As illustrated in Fig. \ref{fig:pipeline}, the framework operates by generating initial semantic proposals and subsequently rectifying them through forensic analysis to achieve both reliable detection and precise localization with Segment Anything Model (SAM) \cite{kirillov2023segment}.

The pipeline begins with a forensics-adapted MLLM that serves as the proposal generator. We fine-tune this model to specialize in forensic analysis tasks, enabling it to generate initial manipulation proposals through semantic understanding and contextual reasoning. Following established practices, we introduce two special tokens <ANL> and <SEG> during processing, where their final hidden layer embeddings, $e^0_{ANL}$ and $e^0_{SEG}$, provide the foundation for subsequent rectification and segmentation processes.

While the MLLM provides high-level semantic analysis, robust forensic detection requires complementary low-level technical evidence. To address this need, we design a Multi-Features Extractor that systematically combines multiple established forensic filters. The extractor integrates SRM, Bayar, Sobel, and Noiseprint++ to generate a unified feature map $F$ through concatenation. This comprehensive forensic representation captures diverse manipulation artifacts including compression inconsistencies, edge discontinuities, and noise pattern irregularities that serve as technical evidence for subsequent tampering validation.

Building upon these initial proposals and extracted forensic features, the Forensics Rectification Module systematically refines the semantic understanding through multi-scale forensic analysis. To enable effective alignment between MLLM analysis and forensic features, we first incorporate a semantic-forensic bridge (Sem-Frs Bridge in Fig.~\ref{fig:pipeline}) that establishes correspondence between high-level semantic understanding and low-level forensic evidence, then selectively emphasize forensic features most appropriate for different analytical scales. Subsequently, both $e^0_{ANL}$ and $e^0_{SEG}$ undergo progressive rectification through multi-scale validation using corresponding forensic evidence at different scales. This process ensures that the rectified output generates reliable detection answers (tampered/real), while the refined segmentation embedding $\hat{e}_{SEG}$ produces more precise guidance for mask decoding.

To further enhance localization precision, we integrate SAM with our Enhanced Segmentation Module, specifically targeting challenging scenarios where tampering operations like inpainting lack explicit visual differences or obvious object boundaries. The module amplifies discrepancies between the extracted forensic features and semantic-focused encoded image embeddings. Through spatial and channel-wise enhancement mechanisms, the module enhances the image embeddings to emphasize forensically significant regions that may be semantically subtle, enabling more accurate forgery localization in challenging manipulation cases.

\subsection{MLLM Proposal Generator}
Multimodal large language models have revolutionized visual understanding by combining language processing with image comprehension. Recent advances, such as Fakeshield \cite{xu2024fakeshield} and ForgeryGPT \cite{liu2024forgerygpt} have demonstrated their potential for forensic applications, where MLLMs can be adapted to identify manipulated images and localize tampered regions by learning to recognize contextual inconsistencies and visual anomalies through text descriptions. However, these approaches face fundamental limitations: MLLMs rely primarily on semantic understanding, making them unreliable for detecting subtle technical artifacts that require low-level forensic analysis, and are prone to hallucinations when making definitive forensic judgments.

Rather than relying on MLLMs as final decision makers, we propose a novel paradigm where they serve as proposal generators that initiate the detection process. We build our proposal generator upon LLaVA \cite{liu2023visual}, leveraging its proven multimodal reasoning capabilities while efficiently adapting it for forensic tasks via Low-Rank Adaptation (LoRA) \cite{hu2022lora, kong2024moe, kong2024open}. Drawing inspiration from LISA \cite{lai2024lisa}, we expand the vocabulary with two special tokens to infuse new forensic abilities: <ANL> for analysis and <SEG> for localization. Given image $I$ and text prompt $T$, the last hidden layer embeddings corresponding to these special tokens are extracted as $e^0_{ANL}$ and $e^0_{SEG}$:
\begin{equation}
[e^0_{ANL}, e^0_{SEG}] = \mathcal{H}(\text{MLLM}(I, T), [pos_{ANL}, pos_{SEG}]),
\end{equation}
where $\mathcal{H}$ represents extracting the last hidden layer of MLLM, and $pos_{ANL}$ and $pos_{SEG}$ indicate the positions of corresponding special tokens.

Crucially, unlike previous approaches that directly provide answers and tampered region masks based solely on MLLM decisions, we treat these embeddings as preliminary proposals requiring validation. The proposal embedding $e^0_{ANL}$ aims to capture the model's initial forensic understanding and reasoning about potential tampering, while $e^0_{SEG}$ provides spatial guidance for localization. This design enables the MLLM to function as a forensic expert that generates informed hypotheses rather than making final judgments. The subsequent rectification process validates and refines these proposals using concrete forensic evidences, effectively combining semantic understanding with technical analysis for robust tampering detection.

\subsection{Multi-Features Extractor}
While high-level analysis provides valuable insights into potential manipulations, robust forensic detection requires technical evidence that extends beyond RGB-domain examination. Sophisticated manipulation techniques can maintain visual plausibility while introducing subtle artifacts detectable only through specialized forensic tools that captures statistical irregularities, compression inconsistencies, and noise pattern disruptions.

Current forensic methods typically employ specific feature types for targeted detection scenarios. RGB-N \cite{zhou2018learning} leverages Bayar filters \cite{bayar2016deep} for compression artifact analysis and JPEG-ComNet \cite{rao2021self} utilizes SRM filters \cite{zhou2018learning} to capture statistical resampling traces. Recognizing the limitations of single-feature approaches, recent methods have explored multi-feature integration to enhance robustness. ManTra-Net \cite{wu2019mantra} and SPAN \cite{hu2020span} combine SRM and Bayar to capture both statistical and compression artifacts, while MVSS-Net \cite{dong2022mvss} fuses Sobel edge features with Bayar for comprehensive boundary analysis.

Building on these insights, we propose a Multi-Features Extractor that systematically integrates four complementary forensic techniques to handle diverse manipulation scenarios with enhanced robustness. As detailed in Table \ref{table:filters}, our approach combines SRM, Bayar, Sobel, and Np++ from TruFor \cite{guillaro2023trufor}, each targeting distinct manipulation signatures and has disparate
advantages and limitations. To jointly utilize them in the following steps, the extracted features are concatenated to form a unified forensic representation
\begin{equation}
F = \text{Concat}([f_{\text{SRM}}(I), f_{\text{Bayar}}(I), f_{\text{Sobel}}(I), f_{\text{Np++}}(I)]),
\end{equation}
where $F \in \mathbb{R}^{H \times W \times K}$ provides $K$ dimensional forensics evidences from different extractors for subsequent refinement processes.

% \begin{table}[t]
%   \centering
%   \caption{Comparative analysis of forensic feature extractor characteristics}
%     \begin{tabular}{>{\raggedright\arraybackslash}p{4em}%
%                     >{\raggedright\arraybackslash}p{7em}%
%                     >{\raggedright\arraybackslash}p{7em}%
%                     >{\raggedright\arraybackslash}p{7em}}
%     \hline
%     Extractor & Captures & Strengths & Limitations \\ \hline
%     Bayar & Constrained CNN; High-order noise residuals. & Suppress image content; Highlights residual artifacts. & Needs training on diverse data. \\ \hline
%     Sobel & Handcrafted for edges and boundaries. & Reveals structural anomalies and seams. & Not noise-specific; Cannot isolate camera/process traces. \\ \hline
%     SRM   & Handcrafted rich model residuals  & Wide coverage of noise patterns. No learned parameters. & Fixed filters may miss unseen artifacts. \\ \hline
%     Noiseprint++ & Learned fingerprint for camera/model pipeline traces and out-camera editing history. & Extracts camera and post-process artifacts; resilient to resizing and compression; captures global anomalies. & Dependence on contrastive fingerprint of seen camera models and editing pipelines. \\ \hline
%     \end{tabular}%
%     \vspace{-3mm}
%   \label{table:filters_txt}%
% \end{table}%

\begin{table}[h!]
    \centering
    \caption{Comparative analysis of forensic feature extractor characteristics}
    \includegraphics[width=1\linewidth]{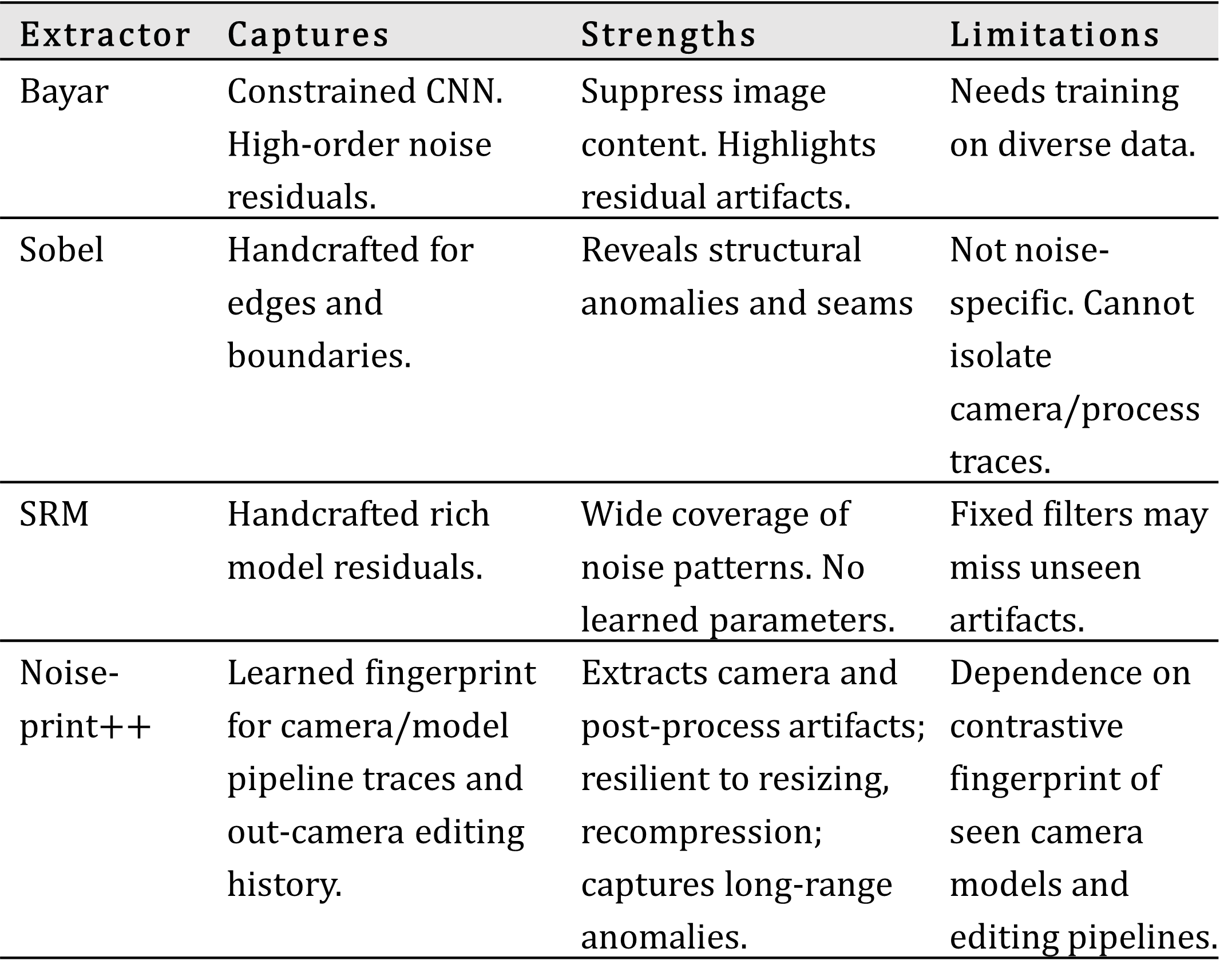}
    \label{table:filters}
\end{table}

\subsection{Forensics Rectification Module}

While MLLMs excel at identifying semantic inconsistencies, their high-level understanding may overlook subtle technical artifacts that constitute definitive forensic evidence. Semantic reasoning alone can produce false positives or miss sophisticated manipulations that preserve contextual coherence while introducing detectable low-level distortions. Previous approaches have demonstrated the value of multi-modal integration, leveraging both handcrafted filters and learned noise patterns to achieve more accurate and robust detection results. These strategies show that different forensic modalities provide complementary perspectives on manipulation artifacts.

\begin{figure}[t]
    \centering
    \vspace{-5mm}
    \includegraphics[width=0.65\linewidth]{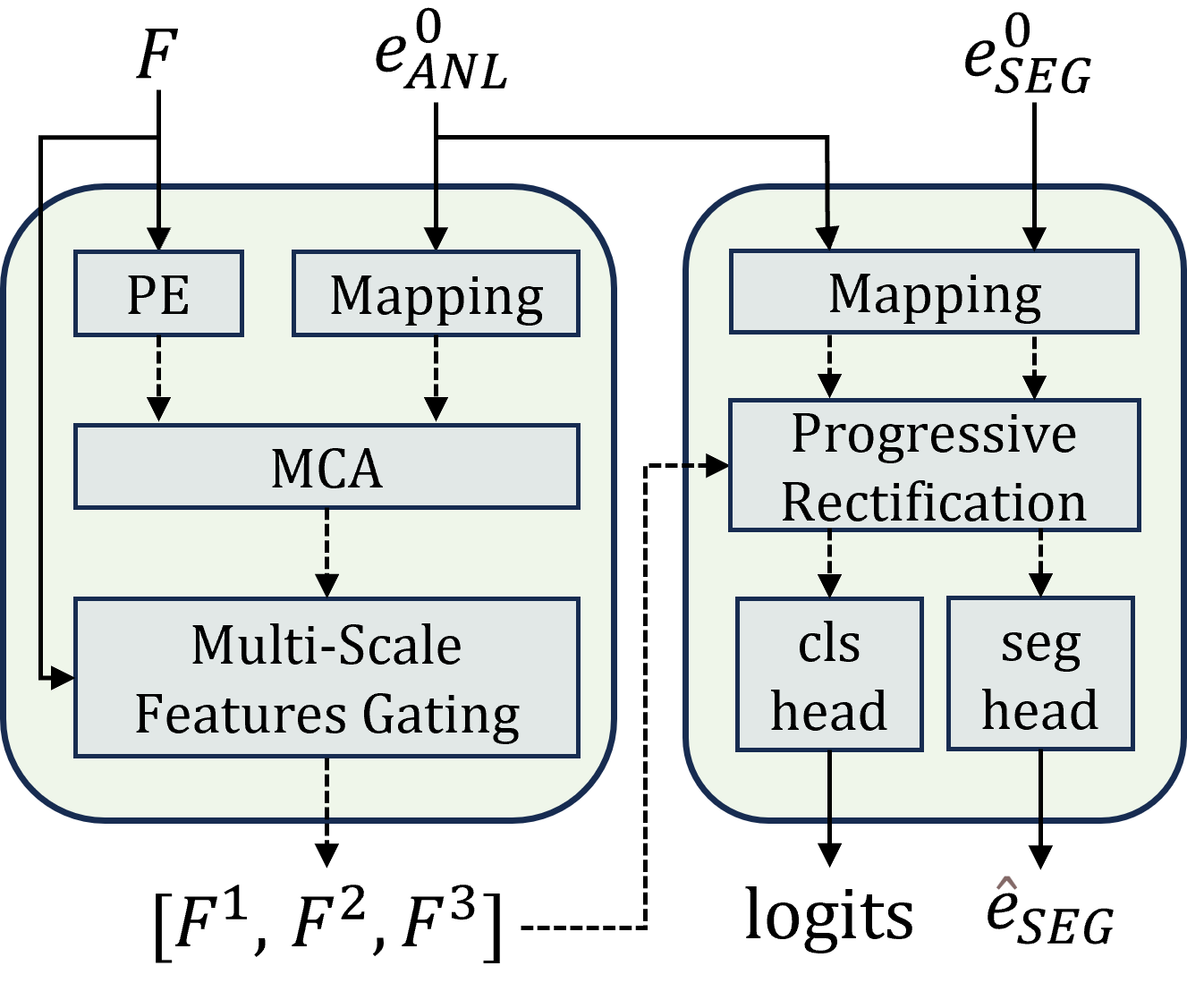}
    \vspace{-2mm}
    \caption{Illustration of the Forensics Rectification Module. (a) Left: Analysis-Informed Feature Gating; (b) Right: Multi-Scale Forensic Rectification.}
    \vspace{-5mm}
    \label{fig:rectify}
\end{figure}

Inspired by this observation, we introduce the Forensics Rectification Module as shown in Fig. \ref{fig:rectify} to systematically verify and refine MLLM proposal embeddings through rigorous examination of low-level forensic evidence. Our approach transcends uniform feature fusion by implementing context-sensitive multi-scale forensic analysis, where the framework dynamically determines optimal analytical strategies based on the primary assessment.

\textbf{Analysis-Informed Feature Gating}. To adapt examination techniques according to unique manipulation characteristics, we establish the analysis-informed feature gating mechanism as illustrated in Fig. \ref{fig:rectify} (a). In order to bridge the semantic understanding from MLLM analysis with the low-level forensic evidence, we first build correspondence between them through multi-head cross-attention (MCA). This enables the analysis embedding $e^0_{ANL}$ to query patch-embedded forensic features $F$, generating contextually-aware gating weights that selectively emphasize critical forensic channels while suppressing irrelevant information:
\begin{equation}
[w^1, w^2, w^3] = \Phi(\text{MCA}(e^0_{ANL}, \text{PE}(F))),
\end{equation}
where $\Phi$ represents the sequential linear transformations and activations that produce feature-wise weight distributions for local, medium and global scale respectively, and PE refers to the patch embedding mechanism. The forensic features are then adaptively weighted through sigmoid function, which targets to generate optimal combinations for multi-scale rectification:

\begin{equation}
F^{k} = \sigma(w^{k}) \odot F, \quad k \in {1, 2, 3}.
\end{equation}

\textbf{Multi-Scale Forensic Rectification}. With the gated features tailored for different analytical scales, we conduct systematic forensic validation through progressive multi-scale examination as depicted in Fig. \ref{fig:rectify} (b). Each scale-specific analysis targets distinct manipulation artifacts: local pixel-level inconsistencies through $3 \times 3$ kernels, medium-scale regional distortions via $7 \times 7$ kernels, and global contextual anomalies using $9 \times 9$ kernels with dilation of 2. The gated features $F^{k}$ undergo scale-appropriate convolution operations and multi-head self-attention (MSA) layers to highlight the pixel-level inconsistencies within each detection granularity:

\begin{equation}
\hat{F}^{k} = \text{MSA}(\text{PE}(\text{Conv}_{s_k}(F^{k}))),
\end{equation}
where $s_k$ represents the convolution kernel for each scale analysis respectively. The resulting forensic evidence then rectifies the semantic embeddings through cross attention, ensuring that both analysis and segmentation understanding align with forensic findings:

\begin{equation}
[e_{ANL}^{k}, e_{SEG}^{k}] = \text{MCA}([e_{ANL}^{k-1}, e_{SEG}^{k-1}], \hat{F}^{k}).
\end{equation}

This iterative rectification progressively refines the embeddings across scales, where each step builds upon previous validation while incorporating new scale-specific evidence. The final rectified results integrate comprehensive multi-scale forensic analysis with initial MLLM proposal, enabling the classification head $h_{c}$ and segmentation head $h_{s}$ to produce more robust detection results and segmentation prompt for SAM decoder that benefits from both high-level reasoning and low-level technical evidence:
\begin{equation}
    [logits, \hat{e}_{SEG}] = [h_{c}(e_{ANL}^{3}),h_{s}(e_{SEG}^{3})].
\end{equation}

Consequently, the proposed rectification mechanism contributes to robust forgery detection by eliminating false positives through comprehensive validation, improving localization accuracy through multi-scale evidence integration, and enhancing generalization by aligning high-level semantic understanding with low-level artifacts. 

\vspace{-3mm} 

\subsection{Enhanced Segmentation Module}

\begin{figure}[t]
    \centering
    \vspace{-3mm} 
    \includegraphics[width=0.5\linewidth]{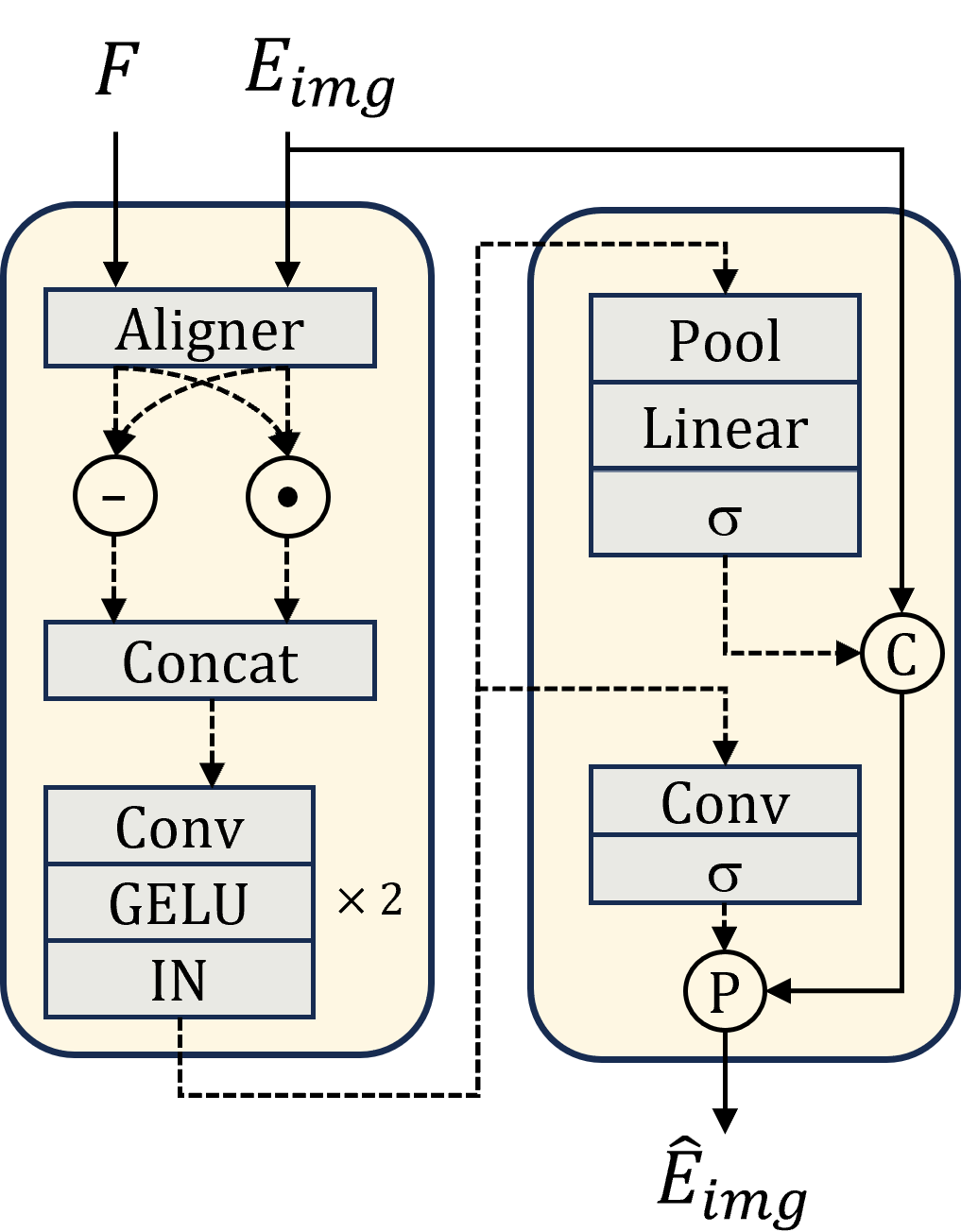}
    \vspace{-2mm} 
    \caption{Diagram of Enhanced Segmentation Module. The discrepancy between encoded image embeddings and forensics features are extracted to enhance the segmentation quality through channel-wise and pixel-wise amplification. IN, Pool and $\sigma$ denote instance normalization, global average pooling and sigmoid function respectively. }
    \vspace{-5mm} 
    \label{fig:amplifier}
\end{figure}

Current MLLM-based segmentation approaches \cite{lai2024lisa, huang2025sida} often integrate with SAM\cite{kirillov2023segment} to leverage its powerful spatial understanding and boundary detection capabilities through the ''embedding-as-mask'' paradigm, where final layer embeddings derived from multimodal reasoning serve as prompts to guide SAM's decoder in generating precise segmentation masks.

Inspired by this success, we initially adopt a similar strategy where segmentation embeddings from our rectified analysis guide SAM's decoder to localize manipulation regions. However, directly employing image embeddings from pretrained encoder presents limitations for accurate forgery localization. SAM's training paradigm focuses on object-level segmentation with semantically coherent boundaries, while tampered regions in sophisticated manipulation may lack explicit semantic boundaries. Advanced techniques specifically aim to create seamless blends that preserve visual continuity across forgery edges, making manipulated content appear contextually plausible in semantic domain.

To overcome this limitation, we propose the Enhanced Segmentation Module as depicted in Fig. \ref{fig:amplifier} which systematically amplifies latent discrepancies between semantic appearance and forensic reality. Given that forensic feature maps $F$ and encoded image embeddings $E_{img}$ typically have different spatial dimensions and channel numbers, we first employ an aligner module with convolutional layers to transform the forensic features to match SAM embeddings' shape, producing aligned forensic features $\tilde{F}$. The module then operates through comprehensive discrepancy extraction that captures multiple aspects of semantic-forensic disagreement. We construct discrepancy embeddings $S$ by concatenating semantic embeddings $E_{img}$, aligned forensic features $\tilde{F}$, their element-wise difference, and their element-wise product:
\begin{equation}
S = \text{Conv}([E_{img}, \tilde{F}, E_{img}-\tilde{F}, E_{img} \odot \tilde{F}]).
\end{equation}

Following discrepancy extraction, we implement adaptive anomaly amplification through dual-gating mechanisms. The spatial component generates pixel-level enhancement weights $S_{p} = \sigma(\text{Conv}(S))$, while channel attention computes global scaling factors $S_{c} = \sigma(\text{MLP}(\text{GAP}(S)))$. Enhanced semantic embeddings are obtained through:

\begin{equation}
\hat{E}_{img} = E_{img} \cdot (1 + S_{c}) \cdot (1 + S_{p}).
\end{equation}

The systematic amplification of semantic-forensic discrepancies transforms imperceptible manipulation signatures into detectable segmentation cues, providing enhanced embeddings necessary for accurate manipulation boundary delineation in challenging scenarios where sophisticated forgeries deliberately obscure traditional segmentation landmarks. Finally, the output mask $M_{pred}$ of forged region is decoded guided by $\hat{e}_{SEG}$ derived from rectification module:
\begin{equation}
    M_{pred}=Decoder(\hat{E}_{img},\hat{e}_{SEG}).
\end{equation}

\subsection{Objective Function}

Our framework is trained end-to-end with a multi-task objective function that jointly optimizes detection accuracy and segmentation precision. The overall training objective combines detection and segmentation losses:
\begin{equation}
\mathcal{L} = \mathcal{L}_{det} + \lambda_{bce} \mathcal{L}_{bce} + \lambda_{dice} \mathcal{L}_{dice},
\end{equation}
where $\mathcal{L}_{det}$ is the cross-entropy loss for classification that distinguishes between authentic and manipulated images. $\mathcal{L}_{bce}$ represents the binary cross-entropy loss for pixel-wise segmentation on tampered regions. $\mathcal{L}_{dice}$ denotes the dice loss that promotes spatial coherence and region-level overlap between predicted and ground truth masks, particularly effective for handling class imbalance in segmentation tasks. The hyperparameters $\lambda_{bce}$ and $\lambda_{dice}$ control the relative importance of the loss components.

\section{Experimental Results}
In this section, we begin by presenting the experimental settings in evaluation. Subsequently, we assess our model's performance through cross-dataset evaluation to examine generalization capabilities and robustness evaluation to test resilience against various perturbations. We further provide qualitative analysis through visualization of manipulation localization results and conduct thorough ablation studies to validate the contribution of each proposed component.

\subsection{Datasets}
In this work, we conduct comprehensive experiments on several widely-used image manipulation datasets. Following established protocols, we utilize CASIAv2 \cite{dong2013casia} as our primary training dataset, which contains 7,491 authentic images and 5,123 manipulated images, providing a substantial foundation for model learning. The remaining ten datasets, including DEF-12k \cite{mahfoudi2019defacto}, Columbia \cite{hsu2006columbia}, IFC \cite{ieee_ifs_tc_2013}, CASIAv1+ \cite{dong2013casia}, COVER \cite{wen2016coverage}, NIST2016 \cite{guan2019mfc}, Carvalho \cite{carvalho2015illuminant}, Korus \cite{korus2016evaluation}, In-the-wild \cite{huh2018fighting}, and IMD2020 \cite{novozamsky2020imd2020}, serve as evaluation benchmarks to assess cross-dataset generalization capabilities. During training, we apply the data augmentation strategy employed in our previous work \cite{kong2025pixel}, which use random perturbations to generate self-blended samples that are semantically impeccable but contain pixel-level inconsistencies, thereby forcing the forensics module to learn more discriminative features for detecting subtle manipulation artifacts and improving overall detection performance.

\begin{table*}[t]
  \centering
  \caption{Pixel-level manipulation localization performance (F1 score)}
    \begin{tabular}{ccccccccccccc}
    \toprule
    Method & Venue & NIST & Columbia & CASIAv1+ & COVER & DEF-12k & IMD & Carvalho & IFC & IntheWild & Korus & Avg \\
    \midrule
    FCN   & CVPR15 & 0.167  & 0.223  & 0.441  & 0.199  & 0.130  & 0.210  & 0.068  & 0.079  & 0.192  & 0.122  & 0.183  \\
    U-Net & MICCAI15 & 0.173  & 0.152  & 0.249  & 0.107  & 0.045  & 0.148  & 0.124  & 0.070  & 0.175  & 0.117  & 0.136  \\
    DeepLabv3 & TPAMI18 & 0.237  & 0.442  & 0.429  & 0.151  & 0.068  & 0.216  & 0.164  & 0.081  & 0.220  & 0.120  & 0.213  \\
    MFCN  & JVCIP18 & 0.243  & 0.184  & 0.346  & 0.148  & 0.067  & 0.170  & 0.150  & 0.098  & 0.161  & 0.118  & 0.169  \\
    RRU-Net & CVPRW19 & 0.200  & 0.264  & 0.291  & 0.078  & 0.033  & 0.159  & 0.084  & 0.052  & 0.178  & 0.097  & 0.144  \\
    MantraNet & CVPR19 & 0.158  & 0.452  & 0.187  & 0.236  & 0.067  & 0.164  & 0.255  & 0.117  & 0.314  & 0.110  & 0.206  \\
    HPFCN  & ICCV19 & 0.172  & 0.115  & 0.173  & 0.104  & 0.038  & 0.111  & 0.082  & 0.065  & 0.125  & 0.097  & 0.108  \\
    H-LSTM & TIP19 & 0.357  & 0.149  & 0.156  & 0.163  & 0.059  & 0.202  & 0.142  & 0.074  & 0.173  & 0.143  & 0.162  \\
    SPAN  & ECCV20 & 0.211  & 0.503  & 0.143  & 0.144  & 0.036  & 0.145  & 0.082  & 0.056  & 0.196  & 0.086  & 0.160  \\
    ViT-B & ICLR21 & 0.254  & 0.217  & 0.282  & 0.142  & 0.062  & 0.154  & 0.169  & 0.071  & 0.208  & 0.176  & 0.174  \\
    Swin-ViT  & ICCV21 & 0.220  & 0.365  & 0.390  & 0.168  & 0.157  & 0.300  & 0.183  & 0.102  & 0.265  & 0.134  & 0.228  \\
    PSCC  & TCSVT22 & 0.173  & 0.503  & 0.335  & 0.220  & 0.072  & 0.197  & \underline{0.295}  & 0.114  & 0.303  & 0.114  & 0.233  \\
    MVSS-Net++ & TPAMI22 & 0.304  & 0.660  & 0.513  & \textbf{0.482}  & 0.095  & 0.270  & 0.271  & 0.080  & 0.295  & 0.102  & 0.307  \\
    CAT-NET & IJCV22 & 0.102  & 0.206  & 0.237  & 0.210  & \textbf{0.206}  & 0.257  & 0.175  & 0.099  & 0.217  & 0.085  & 0.179  \\
    EVP   & CVPR23 & 0.210  & 0.277  & 0.483  & 0.114  & 0.090  & 0.233  & 0.060  & 0.081  & 0.231  & 0.113  & 0.189  \\
    TruFor & CVPR23 & 0.268  & \textbf{0.829}  & 0.532  & 0.280  & 0.148  & 0.359  & 0.213  & 0.127  & 0.361  & 0.122  & 0.324  \\
    PIM   & TPAMI25 & 0.280  & 0.680  & \textbf{0.566}  & 0.251  & 0.167  & \underline{0.419}  & 0.253  & \textbf{0.155}  & \underline{0.418}  & \underline{0.234}  & \underline{0.342}  \\
    SIDA  & CVPR25 & \underline{0.387} & 0.395 & 0.425 & 0.115 & 0.168 & 0.289 & 0.082 & 0.070  & 0.358 & 0.149 & 0.244  \\ \hline
    \rowcolor{gray!20}
    \textbf{PR}    & \textbf{Ours}  & \textbf{0.565} & \underline{0.754} & \underline{0.555} & \underline{0.385} & \underline{0.194} & \textbf{0.488} & \textbf{0.332} & \underline{0.145} & \textbf{0.559} & \textbf{0.253} & \textbf{0.423} \\ \hline
    \end{tabular}%
  \label{tab:pixel_f1}%
\end{table*}%

\begin{table*}[t]
  \centering
  \caption{Pixel-level manipulation localization performance (IoU score)}
    \begin{tabular}{ccccccccccccc}
    \toprule
    Method & Venue & NIST & Columbia & CASIAv1+ & COVER & DEF-12k & IMD & Carvalho & IFC & IntheWild & Korus & Avg \\
    \midrule
    FCN   & CVPR15 & 0.114  & 0.177  & 0.367  & 0.117  & 0.089  & 0.158  & 0.043  & 0.058  & 0.140  & 0.089  & 0.135  \\
    U-Net  & MICCAI15 & 0.128  & 0.097  & 0.204  & 0.072  & 0.031  & 0.105  & 0.082  & 0.048  & 0.121  & 0.082  & 0.097  \\
    DeepLabv3 & TPAMI18 & 0.191  & 0.353  & 0.361  & 0.106  & 0.050  & 0.159  & 0.112  & 0.058  & 0.162  & 0.084  & 0.164  \\
    MFCN  & JVCIP18 & 0.193  & 0.123  & 0.291  & 0.100  & 0.050  & 0.124  & 0.103  & 0.074  & 0.112  & 0.083  & 0.125  \\
    RRU-Net & CVPRW19 & 0.156  & 0.196  & 0.244  & 0.057  & 0.024  & 0.119  & 0.057  & 0.039  & 0.131  & 0.068  & 0.109  \\
    MantraNet & CVPR19 & 0.098  & 0.301  & 0.111  & 0.139  & 0.039  & 0.098  & 0.153  & 0.068  & 0.201  & 0.061  & 0.127  \\
    HPFCN  & ICCV19 & 0.126  & 0.076  & 0.137  & 0.070  & 0.026  & 0.076  & 0.054  & 0.045  & 0.084  & 0.064  & 0.076  \\
    H-LSTM & TIP19 & 0.276  & 0.090  & 0.101  & 0.108  & 0.037  & 0.131  & 0.084  & 0.047  & 0.106  & 0.094  & 0.107  \\
    SPAN  & ECCV20 & 0.156  & 0.390  & 0.112  & 0.105  & 0.024  & 0.100  & 0.049  & 0.037  & 0.132  & 0.055  & 0.116  \\
    ViT-B & ICLR21 & 0.197  & 0.164  & 0.232  & 0.101  & 0.045  & 0.192  & 0.121  & 0.051  & 0.152  & 0.130  & 0.139  \\
    Swin-ViT & ICCV21 & 0.167  & 0.297  & 0.356  & 0.124  & 0.129  & 0.243  & 0.132  & 0.078  & 0.214  & 0.103  & 0.184  \\
    PSCC  & TCSVT22 & 0.108  & 0.360  & 0.232  & 0.130  & 0.042  & 0.120  & 0.185  & 0.067  & 0.193  & 0.066  & 0.150  \\
    MVSS-Net++ & TPAMI22 & 0.239  & 0.573  & 0.397  & \textbf{0.384}  & 0.076  & 0.200  & 0.188  & 0.055  & 0.219  & 0.075  & 0.241  \\
    CAT-NET & IJCV22 & 0.062  & 0.140  & 0.165  & 0.141  & \underline{0.152}  & 0.183  & 0.110  & 0.062  & 0.144  & 0.049  & 0.121  \\
    EVP   & CVPR23 & 0.160  & 0.213  & 0.421  & 0.083  & 0.070  & 0.183  & 0.043  & 0.062  & 0.182  & 0.084  & 0.150  \\
    TruFor & CVPR23 & 0.212  & \textbf{0.781}  & 0.481  & 0.215  & 0.121  & 0.297  & 0.159  & 0.100  & 0.303  & 0.095  & 0.276  \\
    PIM   & TPAMI25 & 0.225  & 0.604  & \textbf{0.512}  & 0.188  & 0.133  & \underline{0.340}  & \underline{0.194}  & \textbf{0.119}  & \underline{0.338}  & \underline{0.182}  & \underline{0.284}  \\
    SIDA  & CVPR25 & \underline{0.304} & 0.315 & 0.356 & 0.081 & 0.128 & 0.218 & 0.057 & 0.049 & 0.279 & 0.105 & 0.189  \\ \hline 
    \rowcolor{gray!20}
    \textbf{PR} & \textbf{Ours} & \textbf{0.474} & \underline{0.663}  & \underline{0.490}  & \underline{0.298}  & \textbf{0.155} & \textbf{0.401} & \textbf{0.261} & \underline{0.113}  & \textbf{0.459} & \textbf{0.193} & \textbf{0.351} \\
    \hline
    \end{tabular}%
  \vspace{-5mm}
  \label{tab:pixel_iou}%
\end{table*}%

\subsection{Evaluation Metrics}
In this paper, three primary evaluation metrics F1, IoU and AUC are employed to assess the performance of our image manipulation detection and localization framework. The F1-score provides a balanced measure between precision and recall:
\begin{equation}
F1 = \frac{2 \times \text{Precision} \times \text{Recall}}{\text{Precision} + \text{Recall}} = \frac{2 \times TP}{2 \times TP + FP + FN},
\end{equation}
where $TP$, $FP$, and $FN$ represent true positives, false positives, and false negatives, respectively. For localization performance specifically, the Intersection over Union (IoU) metric measures the overlap between predicted manipulation masks and ground truth annotations:
\begin{equation}
IoU = \frac{\text{Area of Intersection}}{\text{Area of Union}} = \frac{|M_{pred} \cap M_{gt}|}{|M_{pred} \cup M_{gt}|},
\end{equation}
where $M_{pred}$ denotes the predicted manipulation mask and $M_{gt}$ represents the ground truth mask. The Area Under the Curve (AUC) metric evaluates the overall discriminative performance by measuring the area under the Receiver Operating Characteristic (ROC) curve. 

% which plots the true positive rate against the false positive rate across all thresholds:
% \begin{equation}
% AUC = \int_0^1 TPR(FPR^{-1}(t)) , dt,
% \end{equation}
% where $TPR = \frac{TP}{TP + FN}$ is the true positive rate and $FPR = \frac{FP}{FP + TN}$ is the false positive rate at threshold $t$. The F1-score evaluates accuracy at both image-level detection and pixel-level localization, IoU specifically quantifies the spatial accuracy of manipulation region identification, while AUC provides a threshold-independent assessment of the model's overall capability in localization task.

\begin{table*}[t]
\centering
\caption{Image-level manipulation detection performance (F1 score)}
\begin{tabular}{ccccccccccccc}
    \toprule
	Method & Venue & NIST & Columbia & CASIAv1+ & COVER & DEF-12k & IMD & Carvalho & IFC & In-the-Wild & Korus & Avg  \\ \midrule
	FCN & CVPR15 & 0.897 & 0.702 & 0.713 & 0.653 & 0.607 & 0.827 & 0.566 & 0.441 & 0.908 & 0.627 & 0.694 \\ 
	U-Net & MICCAI15 & 0.945 & 0.692 & 0.673 & \underline{0.660} & 0.633 & 0.878 & 0.662 & 0.466 & 0.972 & 0.637 & 0.722  \\ 
	DeepLabv3 & TPAMI18 & 0.939 & 0.724 & 0.746 & \underline{0.660} & 0.626 & 0.867 & 0.646 & 0.441 & 0.974 & 0.61 & 0.723 \\ 
	RRU-Net & CVPRW19 & 0.871 & 0.678 & 0.661 & 0.553 & 0.564 & 0.798 & 0.646 & 0.387 & 0.877 & 0.587 & 0.662  \\ 
	HPFCN & ICCV19 & 0.893 & 0.664 & 0.58 & 0.624 & 0.615 & 0.824 & 0.636 & 0.446 & 0.902 & 0.632 & 0.682\\ 
	ViT-B & ICLR21 & 0.969 & 0.707 & 0.653 & \textbf{0.671} & 0.646 & 0.87 & 0.664 & 0.448 & 0.972 & 0.644 & 0.724 \\ 
	PSCC & TCSVT22 & 0.953 & 0.698 & 0.577 & \underline{0.660} & 0.646 & 0.866 & 0.674 & 0.463 & 0.972 & 0.649 & 0.716\\ 
	MVSS-Net++ & TPAMI22 & 0.831 & 0.735 & 0.758 & 0.659 & 0.646 & 0.863 & 0.613 & 0.472 & 0.953 & 0.613 & 0.714 \\ 
	CAT-NET  & IJCV22 & \underline{0.982} & 0.687 & 0.548 & 0.641 & 0.642 & 0.885 & 0.662 & 0.464 & \textbf{0.992} & \underline{0.668} & 0.717\\ 
	EVP & CVPR23 & 0.878 & 0.623 & 0.746 & 0.569 & 0.563 & 0.813 & 0.554 & 0.418 & 0.828 & 0.573 & 0.657\\ 
	TruFor & CVPR23 & 0.858 & 0.740 & 0.743 & 0.643 & 0.569 & 0.821 & 0.61 & 0.414 & 0.886 & 0.53 & 0.681 \\ 
	PIM & TPAMI25 & \textbf{0.973} & 0.702 & \underline{0.779} & 0.655 & 0.651 & \underline{0.896} & 0.669 & 0.458 & 0.977 & 0.657 & \underline{0.742} \\ 
    SIDA  & CVPR25 & 0.65  & \underline{0.758} & 0.757 & 0.552 & 0.551 & 0.885 & \underline{0.783} & \textbf{0.638} & 0.833 & 0.368 & 0.678  \\ \hline
	\rowcolor{gray!20} \textbf{PR} & \textbf{Ours} & 0.806 & \textbf{0.788} & \textbf{0.811} & 0.650 & \textbf{0.784} & \textbf{0.971} & \textbf{0.833} & \underline{0.609} & \underline{0.963} & \textbf{0.873} & \textbf{0.809} \\ \hline
\end{tabular}
\vspace{-5mm}
\label{table:image_f1}
\end{table*}

\subsection{Baselines}
To establish a comprehensive benchmark, this paper incorporates 18 representative baseline detectors, including both general-purpose architectures and specialized forgery detection methods. The baselines span three general CNN architectures (FCN \cite{long2015fully}, U-Net \cite{ronneberger2015u}, DeepLabv3 \cite{chen2017deeplab}) and two vision transformers (ViT-B \cite{dosovitskiy2020image}, Swin-ViT \cite{liu2021swin}), alongside eleven state-of-the-art image forgery detection models that capture diverse forgery traces from multiple perspectives: boundary artifacts (MFCN \cite{salloum2018image}, MVSS-Net++ \cite{dong2022mvss}), multi-scale features (PSCC \cite{liu2022pscc}, MVSS-Net++ \cite{dong2022mvss}, TruFor \cite{guillaro2023trufor}), high-frequency artifacts (HPFCN \cite{li2019localization}, MVSS-Net++ \cite{dong2022mvss}, MantraNet \cite{wu2019mantra}), compression artifacts (CAT-NET \cite{kwon2022learning}), noise patterns (H-LSTM \cite{bappy2019hybrid}, TruFor \cite{guillaro2023trufor}), attention mechanisms (SPAN \cite{hu2020span}), unified low-level structure detection (EVP \cite{liu2023explicit}), pixel inconsistency (PIM \cite{kong2025pixel}), and a recent MLLM based forgery detector (SIDA \cite{huang2025sida}). To ensure fair and reproducible comparison, all selected baselines meet at least one of three criteria: publicly available training code, identical training protocol using CASIAv2 dataset, or official pretrained weights.

\subsection{Implementation Details}
Our framework is implemented using the PyTorch library and trained on eight NVIDIA RTX 4090 GPUs. To accelerate computation and reduce memory usage, we leverage bfloat16 mixed-precision training. For optimization, we employ the AdamW optimizer with a learning rate of 5e-5 and set momentum parameters, $\beta_1$ and $\beta_2$, to 0.9 and 0.95, respectively. The weighting coefficients for the loss terms in our composite objective function are uniformly set to 1.0. The training regimen consists of 20 epochs with a linear learning rate warmup schedule for the initial 100 training steps to ensure stable convergence. To monitor performance and prevent overfitting, validation is performed every two epochs, with the checkpoint yielding the best validation metrics being saved for final evaluation. For the multimodal component, we select LLaVA as our base MLLM and fine-tune it to meet the specific demands of forgery analysis.

\subsection{Localization Evaluation}
To comprehensively assess the manipulation localization capabilities of our proposed method, we conduct extensive cross-dataset evaluations on 10 challenging forgery datasets. This evaluation protocol is crucial for understanding real-world performance, as it simulates practical scenarios where models encounter unseen data distributions during deployment. We report both F1-score and IoU metrics with a fixed threshold of 0.5 to ensure fair comparison across all methods, since varying thresholds per dataset would not reflect realistic application constraints. In the presented tables, the best-performing scores are highlighted in bold, while the methods' scores are underlined when achieving second-best results for clarity.

Table \ref{tab:pixel_f1} presents the F1-score results across all test datasets. Our proposed method achieves state-of-the-art performance on 5 out of the 10 datasets: NIST, IMD, Carvalho, In-the-Wild, and Korus. It also demonstrates consistently competitive results across all evaluations. Notably, our method achieves a substantial improvement in overall performance, with an average F1-score of 42.3\%, representing a significant advancement over the previous best-performing method, PIM (34.2\%). The method shows particularly strong performance on challenging datasets, setting the highest score on In-the-Wild (55.9\%) and achieving highly competitive results on others like Columbia (75.4\%) and CASIAv1+ (55.5\%), which demonstrates its outstanding generalization capability across diverse manipulation types and image characteristics.

The IoU evaluation results in Table \ref{tab:pixel_iou} further confirm our method's superior localization accuracy. We achieve the highest IoU scores on 6 out of 10 datasets--NIST, DEF-12k, IMD, Carvalho, In-the-Wild, and Korus--with an average IoU of 35.1\%, substantially outperforming the second-best method PIM (28.4\%). These consistent IoU improvements indicate that our method not only accurately detects manipulated regions but also provides precise delineation of their spatial boundaries.

The comparison with SIDA, another recent MLLM-based approach, demonstrates significant advantages of our framework. Our method outperforms SIDA by 17.9\% points in average F1-score (42.3\% vs. 24.4\%) and 16.2\% points in average IoU (35.1\% vs. 18.9\%). These substantial improvements highlight the superiority of combining semantic reasoning with domain-specific forensic validation rather than relying solely on MLLM outputs, enabling more accurate manipulation localization through forensic evidence-based rectification of MLLM proposals.

\vspace{-3mm} 

\begin{figure*}[t]
    \centering
    \vspace{-5mm} 
    \includegraphics[width=1.0\linewidth]{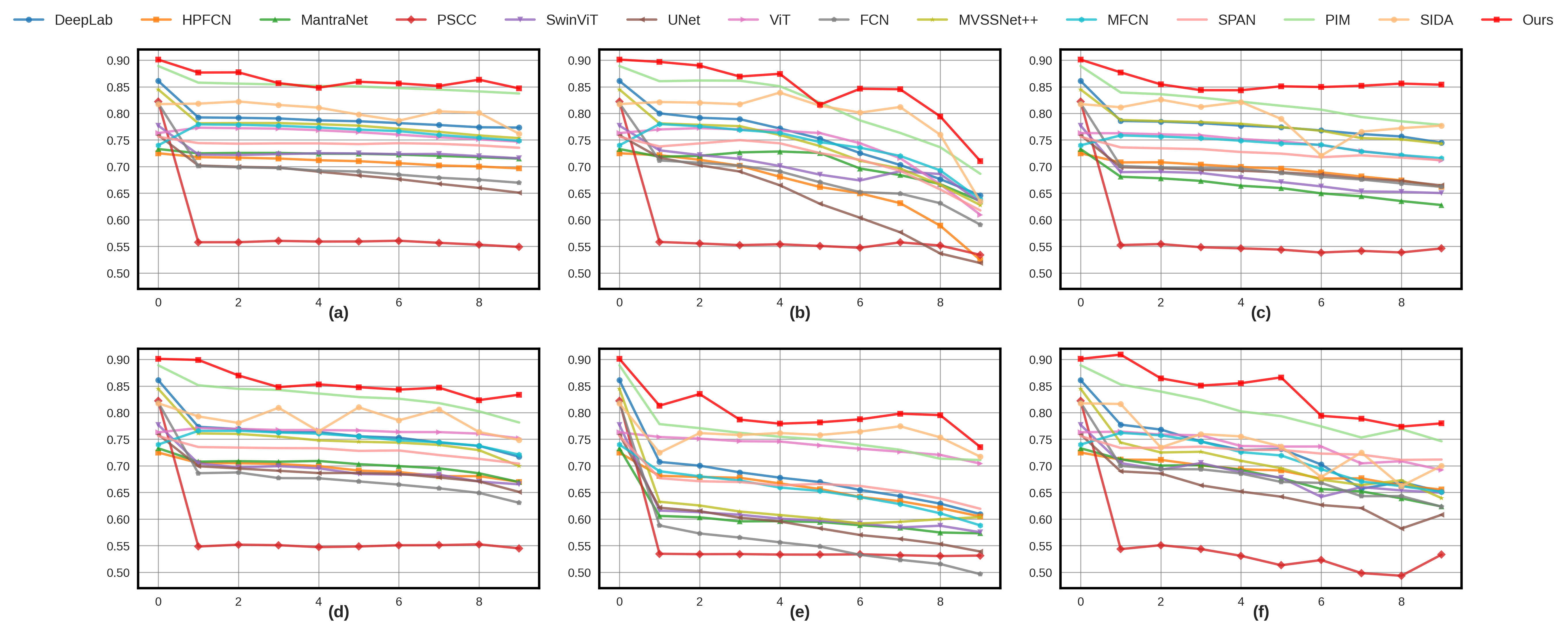}
    \vspace{-3mm} 
    \caption{AUC performance under various image perturbations across increasing severity levels. Evaluation covers six perturbation types: (a) Brightness, (b) Contrast, (c) Darkening, (d) Dithering, (e) JPEG2000 compression, and (f) Pink Noise addition.}
    \vspace{-5mm} 
    \label{fig:perturbation}
\end{figure*}

\subsection{Detection Evaluation}

Beyond pixel-level localization, we evaluate the image-level detection performance to assess our method's capability in distinguishing authentic images from manipulated ones. For image-level detection, our classification head outputs 2-dimensional logits representing authentic and manipulated classes respectively, while baseline methods typically follow the established protocols and use single prediction values with fixed thresholds. 

Table \ref{table:image_f1} presents the image-level F1-score results across all test datasets. Our proposed method demonstrates exceptional detection performance, achieving the highest F1-scores on 6 out of 10 datasets with an average F1-score of 80.9\%. This represents a substantial improvement over existing methods, with our approach outperforming the previous best method PIM (74.2\%) by 6.7 percentage points. Particularly noteworthy are our outstanding results on IMD (97.1\%), Korus (87.3\%), and Carvalho (83.3\%), demonstrating robust detection capability across diverse manipulation scenarios.

The superior image-level performance stems from our framework's ability to effectively bridge semantic understanding with low-level forensic evidence, following a systematic approach where initial hypotheses are validated through technical examination. The MLLM first provides semantic proposals identifying potential inconsistencies, which the Forensics Rectification Module then validates through concrete forensic evidence. This evidence-based validation process ensures that final detection decisions are grounded in technical proof rather than semantic speculation alone, providing more reliable forensic reports. While methods like PIM achieve high performance on specific datasets, our approach maintains more consistent performance across all test scenarios by systematically verifying initial assessments with technical evidence.

\subsection{Robustness Evaluation}
In real-world scenarios, digital images are frequently subjected to various post-processing operations that can significantly interfere with the low-level artifacts many forensic detectors rely on. To assess the resilience of our proposed framework against such common degradations, we conduct a comprehensive robustness evaluation. Our evaluation involves applying six distinct types of perturbations: (a) brightness, (b) contrast, (c) darkening, (d) dithering, (e) JPEG2000 compression, and (f) pink noise—at varying levels of intensity. The results, which plot the localization AUC-score against the perturbation strength, are visualized in Fig. \ref{fig:perturbation}.

As shown in the line charts, our method sustains the highest robustness across all perturbation types and intensities, exhibiting the most gradual decline as degradations increase. We attribute this stability to our Propose-Rectify paradigm: the MLLM's high-level semantic proposal provides a reliable anchor when pixel-level traces are weakened by noise, compression, or tonal shifts. Compared to pixel-level detectors, MLLM-based methods can hold semantic resilience and maintain stronger performance under severe perturbations, underscoring the intrinsic robustness conferred by contextual reasoning.

Building upon this foundation, the meticulous design of our rectification mechanism also contribute to the robustness by refining initial proposals in concrete forensic evidence through adaptive feature gating that emphasize complementary forensic cues into an optimal combination under different perturbation scenarios. This process effectively mitigates erroneous judgment while highlighting artifacts that remain useful and detectable despite image degradations. The synergy between semantic guidance and evidence-driven rectification creates a mutually reinforcing framework: semantic understanding provides robust high-level context that guides forensic analysis, while detailed multi-scale validation by optimal features prevents overreliance on semantic interpretations or single forensics criterion. This dual-layer approach delivers superior and resilient performance across all tested perturbations, demonstrating the effectiveness of combining multimodal reasoning with adaptive forensic verification.

\vspace{-2mm} 

\subsection{Qualitative Experimental Results}

\begin{figure}[t]
    \centering
    \includegraphics[width=1\linewidth]{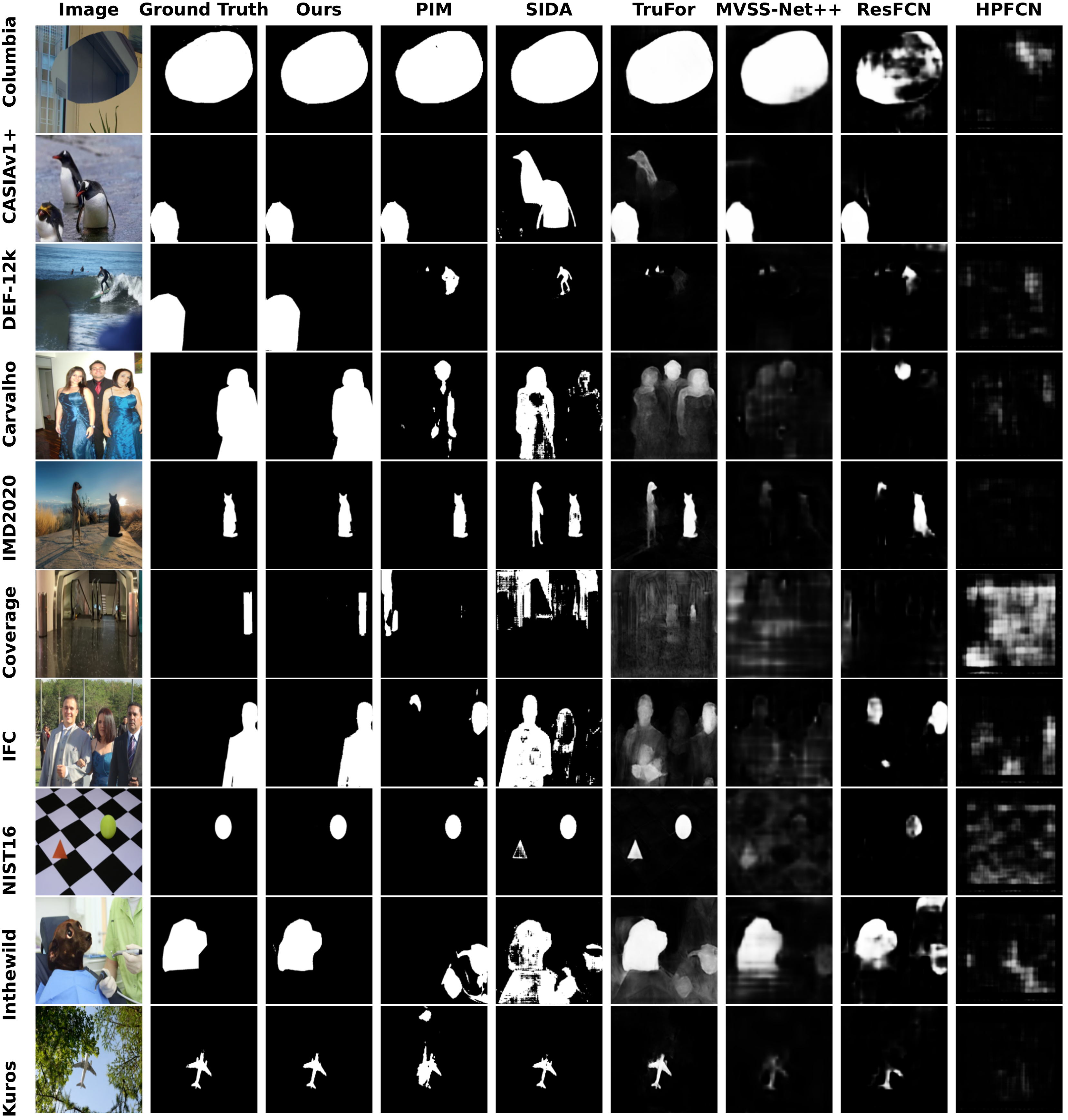}
    \caption{Qualitative comparison of image manipulation localization results. }
    % The leftmost columns show tampered images with ground-truth masks. Remaining columns display predicted masks from state-of-the-art methods and our approach.
    \vspace{-5mm} 
    \label{fig:qualitative}
\end{figure}
To visualize the performance of our framework on image manipulation localization, we present the predicted masks across 10 test datasets in Fig. \ref{fig:qualitative}, comparing our method against multiple baselines. Our proposed method demonstrates consistently superior localization accuracy across diverse manipulation scenarios, achieving the highest precision in identifying tampered regions while maintaining significantly lower false positive rates. Notably, our approach produces remarkably smooth boundaries, particularly evident in challenging datasets like Carvalho and In-the-wild where seamless splicing operations are present. While traditional methods like MVSS-Net++, TruFor, and PIM often generate fragmented or noisy outputs with scattered false positives throughout authentic regions, our paradigm effectively produce clean, coherent masks that closely match ground truth annotations.

The qualitative results highlight several key advantages of our framework over existing approaches. Compared to other MLLM-based methods like SIDA, our approach demonstrates superior precision and consistency through systematic rectification rather than relying solely on multimodal reasoning. Traditional forensic detectors such as ResFCN and HPFCN frequently struggle with boundary localization in challenging scenarios, producing either over-segmented results on complex backgrounds or under-segmented outputs for subtle manipulations. Our Enhanced Segmentation Module effectively addresses these limitations by amplifying semantic-forensic discrepancies before decoding, enabling precise delineation even when manipulation boundaries lack clear objective markers. The consistent performance across varying image qualities, shot environments, and manipulation types validates the robustness of our approach, where MLLM analysis provides stable high-level guidance while low-level forensic rectification ensures accuracy grounded in concrete evidence.

\vspace{-3mm}
\subsection{Ablation Study}

To validate the effectiveness of each component, we conduct comprehensive ablation studies on both detection and localization tasks. We systematically ablate key modules: the MLLM Proposal Generator (PG), Forensics Rectification Module (FRM), Analysis-Informed Feature Gating (FG), and Enhanced Segmentation Module (ESM). The average results across all test datasets are presented in Table \ref{tab:ablation_det} for detection and Table \ref{tab:ablation_loc} for localization.

\begin{table}[t]
    \centering 
    \caption{Ablation study on manipulation localization}
    \begin{tabular}{ccc} 
    \toprule
         Method & AVG. F1 & AVG. IoU \\ \hline\hline
         w/o FRM & 0.301 & 0.225 \\
         w/o FG & 0.360 & 0.292 \\
         w/o PG & 0.367 & 0.297 \\
         w/o ESM & 0.403 & 0.334 \\ \hline
         \rowcolor{gray!20} Ours & \textbf{0.423} & \textbf{0.351} \\
         \hline
    \end{tabular}
    \vspace{-2mm} 
    \label{tab:ablation_loc}
\end{table}

\begin{table}[t]
    \centering 
    \caption{Ablation study on manipulation detection}
    \begin{tabular}{ccc} 
    \toprule
         Method & AVG. ACC & AVG. F1 \\ \hline\hline
         w/o PG & 0.592 & 0.714 \\
         w/o FRM & 0.600 & 0.676 \\
         w/o FG & 0.662 & 0.750 \\ \hline
         \rowcolor{gray!20} Ours & \textbf{0.717} & \textbf{0.809} \\
         \hline
    \end{tabular}
    \vspace{-2mm} 
    \label{tab:ablation_det}
\end{table}

\textbf{Forensics Rectification Module (FRM):} Removing FRM causes the most significant performance degradation. For detection, the average F1-score drops from 0.809 to 0.676, while localization performance falls dramatically from 0.423 to 0.301 (F1) and 0.351 to 0.225 (IoU). This configuration relies solely on MLLM proposals, demonstrating that rectification is the cornerstone of our framework's success. It underscores the criticality of bridging high-level semantic reasoning with fine-grained forensic analysis to correct potential MLLM hallucinations and ground decisions in concrete technical evidence.

\textbf{Analysis-Informed Feature Gating (FG):} Deactivating this mechanism forces the FRM to perform rectification using static, unweighted forensic features, resulting in notable drops to 0.750 (detection F1) and 0.360 (localization F1). This demonstrates that intelligently selecting and emphasizing the most relevant forensic traces based on content analysis, as inferred by the MLLM, is superior to a one-size-fits-all approach and validates our context-sensitive feature gating design.

\textbf{MLLM Proposal Generator (PG):} Operating solely on forensic features without semantic analysis reduces detection F1 to 0.714 and localization F1 to 0.367. This highlights the synergistic relationship between the two pipeline stages. While forensic features are powerful, they benefit immensely from contextual priming provided by the MLLM, which can robustly direct the focus of potentially tampered regions. This confirms the value of our Propose-Rectify paradigm, where semantic proposals effectively guide subsequent validation.

\textbf{Enhanced Segmentation Module (ESM):} Removing this module decreases localization performance to 0.403 (F1) and 0.334 (IoU), confirming its value in achieving precise forgery boundary delineation. This affirms the module's effectiveness in overcoming encoder's inherent bias. By explicitly amplifying discrepancies between forensic evidence and semantic appearance, the ESM enables more accurate segmentation of manipulated regions, especially in challenging cases like inpainting or seamless splicing where clear object boundaries are absent.

In summary, the ablation studies quantitatively validate our core design principles. Each component, from initial semantic proposal and forensic rectification to adaptive feature gating and segmentation enhancement, plays an indispensable and synergistic role, providing evidence that the Propose-Rectify pipeline systematically combining MLLM reasoning with specialized forensic analysis achieves robust and precise manipulation detection and localization.

\section{CONCLUSION AND FUTURE WORK}

In this paper, we introduced a novel Propose-Rectify framework that significantly advances the state-of-the-art in image manipulation detection and localization. We addressed the critical gap between the semantic reasoning capabilities of MLLMs and the necessity for fine-grained, technical forensic analysis. Our framework successfully bridges this divide by first leveraging a forensic-adapted MLLM to propose forgery analysis based on high-level contextual and semantic understanding. Subsequently, our Forensics Rectification Module systematically rectifies these initial proposals using a comprehensive suite of low-level forensic features. Complemented by an Enhanced Segmentation Module that overcomes the inherent biases of SAM encoder, our approach ensures that final outputs are grounded in concrete technical evidences and accurate forgery boundaries. Extensive experimental results across numerous benchmark datasets have unequivocally demonstrated the superiority of our method, showcasing exceptional performance in both detection accuracy and localization precision, as well as robust generalization and resilience to common perturbations.

Despite its strong performance, our framework has certain limitations that present opportunities for future research. The sequential, two-stage nature of the Propose-Rectify pipeline, while effective, introduces greater computational complexity compared to single-stage architectures, which may pose challenges for edge computing applications requiring real-time analysis. To address this computational overhead, several promising directions for future work can be explored. A Mixture-of-Experts approach could be integrated into the MLLM component to selectively activate specialized expert modules based on the detected manipulation type, thereby reducing computational cost while maintaining detection capabilities. Additionally, the core principles of our paradigm are modality-agnostic and hold great potential for extension to other forensic domains, such as video and audio analysis, where semantic inconsistencies can be rectified by analyzing temporal, acoustic, or other domain-specific artifacts. These extensions would further demonstrate the versatility and broad applicability of the Propose-Rectify framework across diverse multimedia forensic scenarios.

\vspace{-2mm}

\ifCLASSOPTIONcaptionsoff
  \newpage
\fi

\bibliographystyle{IEEEtran}
\bibliography{refs}

\begin{thebibliography}{10}
\providecommand{\url}[1]{#1}
\csname url@samestyle\endcsname
\providecommand{\newblock}{\relax}
\providecommand{\bibinfo}[2]{#2}
\providecommand{\BIBentrySTDinterwordspacing}{\spaceskip=0pt\relax}
\providecommand{\BIBentryALTinterwordstretchfactor}{4}
\providecommand{\BIBentryALTinterwordspacing}{\spaceskip=\fontdimen2\font plus
\BIBentryALTinterwordstretchfactor\fontdimen3\font minus \fontdimen4\font\relax}
\providecommand{\BIBforeignlanguage}[2]{{%
\expandafter\ifx\csname l@#1\endcsname\relax
\typeout{** WARNING: IEEEtran.bst: No hyphenation pattern has been}%
\typeout{** loaded for the language `#1'. Using the pattern for}%
\typeout{** the default language instead.}%
\else
\language=\csname l@#1\endcsname
\fi
#2}}
\providecommand{\BIBdecl}{\relax}
\BIBdecl

\bibitem{hochman2014social}
N.~Hochman, ``The social media image,'' \emph{Big Data \& Society}, vol.~1, no.~2, p. 2053951714546645, 2014.

\bibitem{mnookin1998image}
J.~L. Mnookin, ``The image of truth: Photographic evidence and the power of analogy,'' \emph{Yale JL \& Human.}, vol.~10, p.~1, 1998.

\bibitem{kong2022digital}
C.~Kong, S.~Wang, H.~Li \emph{et~al.}, ``Digital and physical face attacks: Reviewing and one step further,'' \emph{APSIPA Transactions on Signal and Information Processing}, vol.~12, no.~1, 2022.

\bibitem{brooks2023instructpix2pix}
T.~Brooks, A.~Holynski, and A.~A. Efros, ``Instructpix2pix: Learning to follow image editing instructions,'' in \emph{Proceedings of the IEEE/CVF conference on computer vision and pattern recognition}, 2023, pp. 18\,392--18\,402.

\bibitem{thakur2020recent}
R.~Thakur and R.~Rohilla, ``Recent advances in digital image manipulation detection techniques: A brief review,'' \emph{Forensic science international}, vol. 312, p. 110311, 2020.

\bibitem{yerushalmy2011digital}
I.~Yerushalmy and H.~Hel-Or, ``Digital image forgery detection based on lens and sensor aberration,'' \emph{International journal of computer vision}, vol.~92, no.~1, pp. 71--91, 2011.

\bibitem{cao2009accurate}
H.~Cao and A.~C. Kot, ``Accurate detection of demosaicing regularity for digital image forensics,'' \emph{IEEE Transactions on Information Forensics and Security}, vol.~4, no.~4, pp. 899--910, 2009.

\bibitem{kwon2022learning}
M.-J. Kwon, S.-H. Nam, I.-J. Yu, H.-K. Lee, and C.~Kim, ``Learning jpeg compression artifacts for image manipulation detection and localization,'' \emph{International Journal of Computer Vision}, vol. 130, no.~8, pp. 1875--1895, 2022.

\bibitem{zhou2018learning}
P.~Zhou, X.~Han, V.~I. Morariu, and L.~S. Davis, ``Learning rich features for image manipulation detection,'' in \emph{Proceedings of the IEEE conference on computer vision and pattern recognition}, 2018, pp. 1053--1061.

\bibitem{li2019localization}
H.~Li and J.~Huang, ``Localization of deep inpainting using high-pass fully convolutional network,'' in \emph{proceedings of the IEEE/CVF international conference on computer vision}, 2019, pp. 8301--8310.

\bibitem{dong2022mvss}
C.~Dong, X.~Chen, R.~Hu, J.~Cao, and X.~Li, ``Mvss-net: Multi-view multi-scale supervised networks for image manipulation detection,'' \emph{IEEE Transactions on Pattern Analysis and Machine Intelligence}, vol.~45, no.~3, pp. 3539--3553, 2022.

\bibitem{wu2019mantra}
Y.~Wu, W.~AbdAlmageed, and P.~Natarajan, ``Mantra-net: Manipulation tracing network for detection and localization of image forgeries with anomalous features,'' in \emph{Proceedings of the IEEE/CVF conference on computer vision and pattern recognition}, 2019, pp. 9543--9552.

\bibitem{xiang2024research}
A.~Xiang, J.~Zhang, Q.~Yang, L.~Wang, and Y.~Cheng, ``Research on splicing image detection algorithms based on natural image statistical characteristics,'' \emph{arXiv preprint arXiv:2404.16296}, 2024.

\bibitem{kong2025pixel}
C.~Kong, A.~Luo, S.~Wang, H.~Li, A.~Rocha, and A.~C. Kot, ``Pixel-inconsistency modeling for image manipulation localization,'' \emph{IEEE Transactions on Pattern Analysis and Machine Intelligence}, 2025.

\bibitem{kong2022detect}
C.~Kong, B.~Chen, H.~Li, S.~Wang, A.~Rocha, and S.~Kwong, ``Detect and locate: Exposing face manipulation by semantic-and noise-level telltales,'' \emph{IEEE Transactions on Information Forensics and Security}, vol.~17, pp. 1741--1756, 2022.

\bibitem{guillaro2023trufor}
F.~Guillaro, D.~Cozzolino, A.~Sud, N.~Dufour, and L.~Verdoliva, ``Trufor: Leveraging all-round clues for trustworthy image forgery detection and localization,'' in \emph{Proceedings of the IEEE/CVF conference on computer vision and pattern recognition}, 2023, pp. 20\,606--20\,615.

\bibitem{luo2010jpeg}
W.~Luo, J.~Huang, and G.~Qiu, ``Jpeg error analysis and its applications to digital image forensics,'' \emph{IEEE Transactions on Information Forensics and Security}, vol.~5, no.~3, pp. 480--491, 2010.

\bibitem{shi2007natural}
Y.~Q. Shi, C.~Chen, and W.~Chen, ``A natural image model approach to splicing detection,'' in \emph{Proceedings of the 9th workshop on Multimedia \& security}, 2007, pp. 51--62.

\bibitem{cozzolino2015splicebuster}
D.~Cozzolino, G.~Poggi, and L.~Verdoliva, ``Splicebuster: A new blind image splicing detector,'' in \emph{2015 IEEE International Workshop on Information Forensics and Security (WIFS)}.\hskip 1em plus 0.5em minus 0.4em\relax IEEE, 2015, pp. 1--6.

\bibitem{fridrich2003detection}
J.~Fridrich, D.~Soukal, J.~Lukas \emph{et~al.}, ``Detection of copy-move forgery in digital images,'' in \emph{Proceedings of digital forensic research workshop}, vol.~3, no.~2.\hskip 1em plus 0.5em minus 0.4em\relax Cleveland, OH, 2003, pp. 652--63.

\bibitem{lin2009fast}
H.-J. Lin, C.-W. Wang, Y.-T. Kao \emph{et~al.}, ``Fast copy-move forgery detection,'' \emph{WSEAS Transactions on Signal Processing}, vol.~5, no.~5, pp. 188--197, 2009.

\bibitem{wu2021iid}
H.~Wu and J.~Zhou, ``Iid-net: Image inpainting detection network via neural architecture search and attention,'' \emph{IEEE Transactions on Circuits and Systems for Video Technology}, vol.~32, no.~3, pp. 1172--1185, 2021.

\bibitem{barglazan2024image}
A.-A. Barglazan, R.~Brad, and C.~Constantinescu, ``Image inpainting forgery detection: a review,'' \emph{Journal of Imaging}, vol.~10, no.~2, p.~42, 2024.

\bibitem{liu2023visual}
H.~Liu, C.~Li, Q.~Wu, and Y.~J. Lee, ``Visual instruction tuning,'' \emph{Advances in neural information processing systems}, vol.~36, pp. 34\,892--34\,916, 2023.

\bibitem{wang2024qwen2}
P.~Wang, S.~Bai, S.~Tan, S.~Wang, Z.~Fan, J.~Bai, K.~Chen, X.~Liu, J.~Wang, W.~Ge \emph{et~al.}, ``Qwen2-vl: Enhancing vision-language model's perception of the world at any resolution,'' \emph{arXiv preprint arXiv:2409.12191}, 2024.

\bibitem{liu2024forgerygpt}
J.~Liu, F.~Zhang, J.~Zhu, E.~Sun, Q.~Zhang, and Z.-J. Zha, ``Forgerygpt: Multimodal large language model for explainable image forgery detection and localization,'' \emph{arXiv preprint arXiv:2410.10238}, 2024.

\bibitem{xu2024fakeshield}
Z.~Xu, X.~Zhang, R.~Li, Z.~Tang, Q.~Huang, and J.~Zhang, ``Fakeshield: Explainable image forgery detection and localization via multi-modal large language models,'' \emph{arXiv preprint arXiv:2410.02761}, 2024.

\bibitem{huang2025sida}
Z.~Huang, J.~Hu, X.~Li, Y.~He, X.~Zhao, B.~Peng, B.~Wu, X.~Huang, and G.~Cheng, ``Sida: Social media image deepfake detection, localization and explanation with large multimodal model,'' in \emph{Proceedings of the Computer Vision and Pattern Recognition Conference}, 2025, pp. 28\,831--28\,841.

\bibitem{yildirim2024task}
I.~Yildirim and L.~Paul, ``From task structures to world models: what do llms know?'' \emph{Trends in Cognitive Sciences}, vol.~28, no.~5, pp. 404--415, 2024.

\bibitem{radford2021learning}
A.~Radford, J.~W. Kim, C.~Hallacy, A.~Ramesh, G.~Goh, S.~Agarwal, G.~Sastry, A.~Askell, P.~Mishkin, J.~Clark \emph{et~al.}, ``Learning transferable visual models from natural language supervision,'' in \emph{International conference on machine learning}.\hskip 1em plus 0.5em minus 0.4em\relax PmLR, 2021, pp. 8748--8763.

\bibitem{huang2024ffaa}
Z.~Huang, B.~Xia, Z.~Lin, Z.~Mou, W.~Yang, and J.~Jia, ``Ffaa: Multimodal large language model based explainable open-world face forgery analysis assistant,'' \emph{arXiv preprint arXiv:2408.10072}, 2024.

\bibitem{achiam2023gpt}
J.~Achiam, S.~Adler, S.~Agarwal, L.~Ahmad, I.~Akkaya, F.~L. Aleman, D.~Almeida, J.~Altenschmidt, S.~Altman, S.~Anadkat \emph{et~al.}, ``Gpt-4 technical report,'' \emph{arXiv preprint arXiv:2303.08774}, 2023.

\bibitem{piva2013overview}
A.~Piva, ``An overview on image forensics,'' \emph{International Scholarly Research Notices}, vol. 2013, no.~1, p. 496701, 2013.

\bibitem{nist}
\BIBentryALTinterwordspacing
Standard guide for image authentication. National Institute of Standards and Technology. [Online]. Available: \url{https://www.nist.gov}
\BIBentrySTDinterwordspacing

\bibitem{ferrara2012image}
P.~Ferrara, T.~Bianchi, A.~De~Rosa, and A.~Piva, ``Image forgery localization via fine-grained analysis of cfa artifacts,'' \emph{IEEE Transactions on Information Forensics and Security}, vol.~7, no.~5, pp. 1566--1577, 2012.

\bibitem{fu2012forgery}
H.~Fu and X.~Cao, ``Forgery authentication in extreme wide-angle lens using distortion cue and fake saliency map,'' \emph{IEEE Transactions on Information Forensics and Security}, vol.~7, no.~4, pp. 1301--1314, 2012.

\bibitem{gloe2010efficient}
T.~Gloe, K.~Borowka, and A.~Winkler, ``Efficient estimation and large-scale evaluation of lateral chromatic aberration for digital image forensics,'' in \emph{Media Forensics and Security II}, vol. 7541.\hskip 1em plus 0.5em minus 0.4em\relax SPIE, 2010, pp. 62--74.

\bibitem{kobayashi2010detecting}
M.~Kobayashi, T.~Okabe, and Y.~Sato, ``Detecting forgery from static-scene video based on inconsistency in noise level functions,'' \emph{IEEE Transactions on Information Forensics and Security}, vol.~5, no.~4, pp. 883--892, 2010.

\bibitem{lyu2014exposing}
S.~Lyu, X.~Pan, and X.~Zhang, ``Exposing region splicing forgeries with blind local noise estimation,'' \emph{International journal of computer vision}, vol. 110, no.~2, pp. 202--221, 2014.

\bibitem{bianchi2012image}
T.~Bianchi and A.~Piva, ``Image forgery localization via block-grained analysis of jpeg artifacts,'' \emph{IEEE Transactions on Information Forensics and Security}, vol.~7, no.~3, pp. 1003--1017, 2012.

\bibitem{chen2011detecting}
Y.-L. Chen and C.-T. Hsu, ``Detecting recompression of jpeg images via periodicity analysis of compression artifacts for tampering detection,'' \emph{IEEE Transactions on Information Forensics and Security}, vol.~6, no.~2, pp. 396--406, 2011.

\bibitem{farid2009exposing}
H.~Farid, ``Exposing digital forgeries from jpeg ghosts,'' \emph{IEEE transactions on information forensics and security}, vol.~4, no.~1, pp. 154--160, 2009.

\bibitem{yao2011detecting}
H.~Yao, S.~Wang, Y.~Zhao, and X.~Zhang, ``Detecting image forgery using perspective constraints,'' \emph{IEEE Signal Processing Letters}, vol.~19, no.~3, pp. 123--126, 2011.

\bibitem{peng2016optimized}
B.~Peng, W.~Wang, J.~Dong, and T.~Tan, ``Optimized 3d lighting environment estimation for image forgery detection,'' \emph{IEEE Transactions on Information Forensics and Security}, vol.~12, no.~2, pp. 479--494, 2016.

\bibitem{cozzolino2019noiseprint}
D.~Cozzolino and L.~Verdoliva, ``Noiseprint: A cnn-based camera model fingerprint,'' \emph{IEEE Transactions on Information Forensics and Security}, vol.~15, pp. 144--159, 2019.

\bibitem{rao2022towards}
Y.~Rao, J.~Ni, W.~Zhang, and J.~Huang, ``Towards jpeg-resistant image forgery detection and localization via self-supervised domain adaptation,'' \emph{IEEE transactions on pattern analysis and machine intelligence}, vol.~47, no.~5, pp. 3285--3297, 2022.

\bibitem{wang2022jpeg}
M.~Wang, X.~Fu, J.~Liu, and Z.-J. Zha, ``Jpeg compression-aware image forgery localization,'' in \emph{Proceedings of the 30th ACM International Conference on Multimedia}, 2022, pp. 5871--5879.

\bibitem{hu2020span}
X.~Hu, Z.~Zhang, Z.~Jiang, S.~Chaudhuri, Z.~Yang, and R.~Nevatia, ``Span: Spatial pyramid attention network for image manipulation localization,'' in \emph{European conference on computer vision}.\hskip 1em plus 0.5em minus 0.4em\relax Springer, 2020, pp. 312--328.

\bibitem{rao2021self}
Y.~Rao and J.~Ni, ``Self-supervised domain adaptation for forgery localization of jpeg compressed images,'' in \emph{Proceedings of the IEEE/CVF international conference on computer vision}, 2021, pp. 15\,034--15\,043.

\bibitem{liu2023explicit}
W.~Liu, X.~Shen, C.-M. Pun, and X.~Cun, ``Explicit visual prompting for low-level structure segmentations,'' in \emph{Proceedings of the IEEE/CVF Conference on Computer Vision and Pattern Recognition}, 2023, pp. 19\,434--19\,445.

\bibitem{lai2024lisa}
X.~Lai, Z.~Tian, Y.~Chen, Y.~Li, Y.~Yuan, S.~Liu, and J.~Jia, ``Lisa: Reasoning segmentation via large language model,'' in \emph{Proceedings of the IEEE/CVF Conference on Computer Vision and Pattern Recognition}, 2024, pp. 9579--9589.

\bibitem{liu2025seg}
Y.~Liu, B.~Peng, Z.~Zhong, Z.~Yue, F.~Lu, B.~Yu, and J.~Jia, ``Seg-zero: Reasoning-chain guided segmentation via cognitive reinforcement,'' \emph{arXiv preprint arXiv:2503.06520}, 2025.

\bibitem{guo2025rethinking}
X.~Guo, X.~Song, Y.~Zhang, X.~Liu, and X.~Liu, ``Rethinking vision-language model in face forensics: Multi-modal interpretable forged face detector,'' in \emph{Proceedings of the Computer Vision and Pattern Recognition Conference}, 2025, pp. 105--116.

\bibitem{cheng2025empowering}
F.~Cheng, H.~Li, F.~Liu, R.~van Rooij, K.~Zhang, and Z.~Lin, ``Empowering llms with logical reasoning: A comprehensive survey,'' \emph{arXiv preprint arXiv:2502.15652}, 2025.

\bibitem{wu2024get}
Z.~Wu, M.~Jiang, and C.~Shen, ``Get an a in math: Progressive rectification prompting,'' in \emph{Proceedings of the AAAI Conference on Artificial Intelligence}, vol.~38, no.~17, 2024, pp. 19\,288--19\,296.

\bibitem{guo2025deepseek}
D.~Guo, D.~Yang, H.~Zhang, J.~Song, R.~Zhang, R.~Xu, Q.~Zhu, S.~Ma, P.~Wang, X.~Bi \emph{et~al.}, ``Deepseek-r1: Incentivizing reasoning capability in llms via reinforcement learning,'' \emph{arXiv preprint arXiv:2501.12948}, 2025.

\bibitem{huang2023agentcoder}
D.~Huang, J.~M. Zhang, M.~Luck, Q.~Bu, Y.~Qing, and H.~Cui, ``Agentcoder: Multi-agent-based code generation with iterative testing and optimisation,'' \emph{arXiv preprint arXiv:2312.13010}, 2023.

\bibitem{lei2024spider}
F.~Lei, J.~Chen, Y.~Ye, R.~Cao, D.~Shin, H.~Su, Z.~Suo, H.~Gao, W.~Hu, P.~Yin \emph{et~al.}, ``Spider 2.0: Evaluating language models on real-world enterprise text-to-sql workflows,'' \emph{arXiv preprint arXiv:2411.07763}, 2024.

\bibitem{chen2024internet}
W.~Chen, Z.~You, R.~Li, Y.~Guan, C.~Qian, C.~Zhao, C.~Yang, R.~Xie, Z.~Liu, and M.~Sun, ``Internet of agents: Weaving a web of heterogeneous agents for collaborative intelligence,'' \emph{arXiv preprint arXiv:2407.07061}, 2024.

\bibitem{guo2023hierarchical}
X.~Guo, X.~Liu, Z.~Ren, S.~Grosz, I.~Masi, and X.~Liu, ``Hierarchical fine-grained image forgery detection and localization,'' in \emph{Proceedings of the IEEE/CVF conference on computer vision and pattern recognition}, 2023, pp. 3155--3165.

\bibitem{jia2018coarse}
S.~Jia, Z.~Xu, H.~Wang, C.~Feng, and T.~Wang, ``Coarse-to-fine copy-move forgery detection for video forensics,'' \emph{IEEE Access}, vol.~6, pp. 25\,323--25\,335, 2018.

\bibitem{kirillov2023segment}
A.~Kirillov, E.~Mintun, N.~Ravi, H.~Mao, C.~Rolland, L.~Gustafson, T.~Xiao, S.~Whitehead, A.~C. Berg, W.-Y. Lo \emph{et~al.}, ``Segment anything,'' in \emph{Proceedings of the IEEE/CVF international conference on computer vision}, 2023, pp. 4015--4026.

\bibitem{hu2022lora}
E.~J. Hu, Y.~Shen, P.~Wallis, Z.~Allen-Zhu, Y.~Li, S.~Wang, L.~Wang, W.~Chen \emph{et~al.}, ``Lora: Low-rank adaptation of large language models.'' \emph{ICLR}, vol.~1, no.~2, p.~3, 2022.

\bibitem{kong2024moe}
C.~Kong, A.~Luo, P.~Bao, Y.~Yu, H.~Li, Z.~Zheng, S.~Wang, and A.~C. Kot, ``Moe-ffd: Mixture of experts for generalized and parameter-efficient face forgery detection,'' \emph{arXiv preprint arXiv:2404.08452}, 2024.

\bibitem{kong2024open}
C.~Kong, A.~Luo, P.~Bao, H.~Li, R.~Wan, Z.~Zheng, A.~Rocha, and A.~C. Kot, ``Open-set deepfake detection: a parameter-efficient adaptation method with forgery style mixture,'' \emph{arXiv preprint arXiv:2408.12791}, 2024.

\bibitem{bayar2016deep}
B.~Bayar and M.~C. Stamm, ``A deep learning approach to universal image manipulation detection using a new convolutional layer,'' in \emph{Proceedings of the 4th ACM workshop on information hiding and multimedia security}, 2016, pp. 5--10.

\bibitem{dong2013casia}
J.~Dong, W.~Wang, and T.~Tan, ``Casia image tampering detection evaluation database,'' in \emph{2013 IEEE China summit and international conference on signal and information processing}.\hskip 1em plus 0.5em minus 0.4em\relax IEEE, 2013, pp. 422--426.

\bibitem{mahfoudi2019defacto}
G.~Mahfoudi, B.~Tajini, F.~Retraint, F.~Morain-Nicolier, J.~L. Dugelay, and M.~Pic, ``Defacto: Image and face manipulation dataset,'' in \emph{2019 27Th european signal processing conference (EUSIPCO)}.\hskip 1em plus 0.5em minus 0.4em\relax IEEE, 2019, pp. 1--5.

\bibitem{hsu2006columbia}
J.~Hsu and S.~Chang, ``Columbia uncompressed image splicing detection evaluation dataset,'' \emph{Columbia DVMM Research Lab}, vol.~6, 2006.

\bibitem{ieee_ifs_tc_2013}
``{IEEE IFS-TC} image forensics challenge dataset,'' 2013, [Online]. Available: https://signalprocessingsociety.org/newsletter/2013/06/ifs-tc-image-forensics-challenge.

\bibitem{wen2016coverage}
B.~Wen, Y.~Zhu, R.~Subramanian, T.-T. Ng, X.~Shen, and S.~Winkler, ``Coverage—a novel database for copy-move forgery detection,'' in \emph{2016 IEEE international conference on image processing (ICIP)}.\hskip 1em plus 0.5em minus 0.4em\relax IEEE, 2016, pp. 161--165.

\bibitem{guan2019mfc}
H.~Guan, M.~Kozak, E.~Robertson, Y.~Lee, A.~N. Yates, A.~Delgado, D.~Zhou, T.~Kheyrkhah, J.~Smith, and J.~Fiscus, ``Mfc datasets: Large-scale benchmark datasets for media forensic challenge evaluation,'' in \emph{2019 IEEE Winter Applications of Computer Vision Workshops (WACVW)}.\hskip 1em plus 0.5em minus 0.4em\relax IEEE, 2019, pp. 63--72.

\bibitem{carvalho2015illuminant}
T.~Carvalho, F.~A. Faria, H.~Pedrini, R.~d.~S. Torres, and A.~Rocha, ``Illuminant-based transformed spaces for image forensics,'' \emph{IEEE transactions on information forensics and security}, vol.~11, no.~4, pp. 720--733, 2015.

\bibitem{korus2016evaluation}
P.~Korus and J.~Huang, ``Evaluation of random field models in multi-modal unsupervised tampering localization,'' in \emph{2016 IEEE international workshop on information forensics and security (WIFS)}.\hskip 1em plus 0.5em minus 0.4em\relax IEEE, 2016, pp. 1--6.

\bibitem{huh2018fighting}
M.~Huh, A.~Liu, A.~Owens, and A.~A. Efros, ``Fighting fake news: Image splice detection via learned self-consistency,'' in \emph{Proceedings of the European conference on computer vision (ECCV)}, 2018, pp. 101--117.

\bibitem{novozamsky2020imd2020}
A.~Novozamsky, B.~Mahdian, and S.~Saic, ``Imd2020: A large-scale annotated dataset tailored for detecting manipulated images,'' in \emph{Proceedings of the IEEE/CVF winter conference on applications of computer vision workshops}, 2020, pp. 71--80.

\bibitem{long2015fully}
J.~Long, E.~Shelhamer, and T.~Darrell, ``Fully convolutional networks for semantic segmentation,'' in \emph{Proceedings of the IEEE conference on computer vision and pattern recognition}, 2015, pp. 3431--3440.

\bibitem{ronneberger2015u}
O.~Ronneberger, P.~Fischer, and T.~Brox, ``U-net: Convolutional networks for biomedical image segmentation,'' in \emph{International Conference on Medical image computing and computer-assisted intervention}.\hskip 1em plus 0.5em minus 0.4em\relax Springer, 2015, pp. 234--241.

\bibitem{chen2017deeplab}
L.-C. Chen, G.~Papandreou, I.~Kokkinos, K.~Murphy, and A.~L. Yuille, ``Deeplab: Semantic image segmentation with deep convolutional nets, atrous convolution, and fully connected crfs,'' \emph{IEEE transactions on pattern analysis and machine intelligence}, vol.~40, no.~4, pp. 834--848, 2017.

\bibitem{dosovitskiy2020image}
A.~Dosovitskiy, L.~Beyer, A.~Kolesnikov, D.~Weissenborn, X.~Zhai, T.~Unterthiner, M.~Dehghani, M.~Minderer, G.~Heigold, S.~Gelly \emph{et~al.}, ``An image is worth 16x16 words: Transformers for image recognition at scale,'' \emph{arXiv preprint arXiv:2010.11929}, 2020.

\bibitem{liu2021swin}
Z.~Liu, Y.~Lin, Y.~Cao, H.~Hu, Y.~Wei, Z.~Zhang, S.~Lin, and B.~Guo, ``Swin transformer: Hierarchical vision transformer using shifted windows,'' in \emph{Proceedings of the IEEE/CVF international conference on computer vision}, 2021, pp. 10\,012--10\,022.

\bibitem{salloum2018image}
R.~Salloum, Y.~Ren, and C.-C.~J. Kuo, ``Image splicing localization using a multi-task fully convolutional network (mfcn),'' \emph{Journal of Visual Communication and Image Representation}, vol.~51, pp. 201--209, 2018.

\bibitem{liu2022pscc}
X.~Liu, Y.~Liu, J.~Chen, and X.~Liu, ``Pscc-net: Progressive spatio-channel correlation network for image manipulation detection and localization,'' \emph{IEEE Transactions on Circuits and Systems for Video Technology}, vol.~32, no.~11, pp. 7505--7517, 2022.

\bibitem{bappy2019hybrid}
J.~H. Bappy, C.~Simons, L.~Nataraj, B.~Manjunath, and A.~K. Roy-Chowdhury, ``Hybrid lstm and encoder--decoder architecture for detection of image forgeries,'' \emph{IEEE transactions on image processing}, vol.~28, no.~7, pp. 3286--3300, 2019.

\end{thebibliography}

\end{document}